\documentclass[10pt,journal,cspaper,compsoc]{IEEEtran}

\usepackage{amsmath, amsthm}
\usepackage{eepic, epic,color}
\usepackage{times}
\usepackage[latin1]{inputenc}
\usepackage{amsfonts}
\usepackage{psfrag}
\usepackage{amsbsy}
\usepackage{amssymb}
\usepackage[mathscr]{eucal}
\usepackage{latexsym}
\usepackage[T1]{fontenc}
\usepackage{amsmath}
\usepackage{comment}
\usepackage[dvips]{graphicx}
\usepackage{bm}
\usepackage{array}
\usepackage{verbatim,mathrsfs, cite}
\usepackage{algorithmic}
\usepackage{algorithm}
\usepackage{multirow}

\newcounter{lecnum}
\newtheorem{theorem}{Theorem}
\newtheorem{lemma}{Lemma}

\theoremstyle{remark}
\newtheorem{rem}{Remark}  

\newcommand{\figw}{0.95\columnwidth}

\newcommand{\sgn}{\operatornamewithlimits{sgn}}

\newcommand{\be}{\begin{equation}}
\newcommand{\ee}{\end{equation}}
\newcommand{\ba}{\begin{align}}

\newcommand{\abs}[1]{\lvert#1\rvert}
\newcommand{\norm}[1]{\lVert#1\rVert}

\def\B(#1){\hbox{\mathbf$#1$}}
\def\C(#1){{\cal #1}}

\graphicspath{{./Figure/}}

\begin{document}

\title{Ensemble of Distributed Learners for Online Classification of Dynamic Data Streams}
\author{
Luca~Canzian,~\IEEEmembership{Member,~IEEE,}
Yu~Zhang, 
and~Mihaela~van~der~Schaar,~\IEEEmembership{Fellow,~IEEE}
\IEEEcompsocitemizethanks{\IEEEcompsocthanksitem The authors are with the Department of Electrical Engineering, UCLA, Los Angeles CA 90095, USA.
}
\IEEEcompsocitemizethanks{\IEEEcompsocthanksitem This work was partially supported by the AFOSR DDDAS grant and the NSF CCF 1016081 grant.
}
}

\IEEEcompsoctitleabstractindextext{%
\begin{abstract}
We present an efficient distributed online learning scheme to classify data captured from distributed, heterogeneous, and dynamic data sources.
Our scheme consists of multiple distributed local learners, that analyze different streams of data that are correlated to a common event that needs to be classified.
Each learner uses a local classifier to make a local prediction.
The local predictions are then collected by each learner and combined using a weighted majority rule to output the final prediction.
We propose a novel online ensemble learning algorithm to update the aggregation rule in order to adapt to the underlying data dynamics.
We rigorously determine a bound for the worst--case mis--classification probability of our algorithm which depends on the mis--classification probabilities of the best static aggregation rule, and of the best local classifier. 
Importantly, the worst--case mis--classification probability of our algorithm tends asymptotically to $0$ if the mis--classification probability of the best static aggregation rule or the mis--classification probability of the best local classifier tend to $0$.
Then we extend our algorithm to address challenges specific to the distributed implementation and we prove new bounds that apply to these settings.
Finally, we test our scheme by performing an evaluation study on several data sets. 
When applied to data sets widely used by the literature dealing with dynamic data streams and concept drift, our scheme exhibits performance gains ranging from $34\%$ to $71\%$ with respect to state--of--the--art solutions.
\end{abstract}

\begin{keywords}
Online learning, distributed learning, ensemble of classifiers, dynamic streams, concept drift, classification.
\end{keywords}}

\maketitle

\section{Introduction}
\label{sec:intro} 

Recent years have witnessed the proliferation of data--driven applications that exploit the large amount of data captured from distributed, heterogeneous, and dynamic (i.e., whose characteristics are varying over time) data sources.
Examples of such applications include surveillance \cite{Stauffer00}, driver assistance systems \cite{Avidan04}, network monitoring \cite{Gao2007}, social multimedia \cite{Lin13}, and patient monitoring \cite{Tseng08}. 
However, the effective utilization of such high-volume data also involves significant challenges that are the main concern of this work.
First, the captured data need to be analyzed \emph{online} (e.g., to make predictions and timely decisions based on these predictions); thus, the learning algorithms need to deal with the time--varying characteristics of the underlying data, i.e., adequately deal with \emph{concept--drift} \cite{Zliobaite10}.
Second, the privacy, communication, and sharing costs make it difficult to collect and store all the observed data.
Third, the devices that collect the data may be managed by different entities (e.g., multiple hospitals, multiple camera systems, multiple routers, etc.) and may follow policies (e.g., type of information to exchange, rate at which data are collected, etc.) that are not centrally controllable.

To address these challenges, we propose an \emph{online ensemble learning} technique, which we refer to as Perceptron Weighted Majority (PWM).
Specifically, we consider a set of distributed learners that observe data from different sources, which are correlated to a common event that must be classified by the learners (see Fig. \ref{fig:syst}).
We focus on binary classification problems.\footnote{We remark that a multi--class classifier can be decomposed as a cascade of binary classifiers \cite{Ducasse10}.} 
For each single instance that enters the system, each learner makes the final classification decision by collecting the local predictions of all the learners and combining them using a weighted majority rule as in \cite{Freund1997, Littlestone1994, Littlestone1988, 
Blum1997, Fan1999, Wang2003, 
Masud2009, Herbster1998, Kolter2005, Kolter2007}.
After having made the final prediction, the learner is told the real value, i.e., the \emph{label}, associated to the event to classify.
Exploiting such information, the learner updates the aggregation weights adopting a perceptron learning rule \cite{Rosenblatt1957}.

The main features of our scheme are:

\textbf{DIS: Distributed data streams.}
The majority of the existing ensemble schemes proposed in literature assume that the learners make a prediction after having observed the same data \cite{Fan1999, Street2001, Wang2003, 
Masud2009, Littlestone1994, Littlestone1988, 
Blum1997, Herbster1998, Kolter2005, Kolter2007, 
Minku2012}. 
Our approach does not make such an assumption, allowing for the possibility that the distributed learners observe \emph{different} correlated data streams.
In particular, the statistical dependency among the label and the observation of a learner can be different from the statistical dependency among the label and the observation of another learner, i.e., each source has a specific generating process \cite{Geras13}.

\textbf{DYN: Dynamic data streams.}
Many existing ensemble schemes \cite{Freund1997, Littlestone1994, Littlestone1988, 
Blum1997, Fan1999} assume that the data are generated from a stationary distribution, i.e., that the \emph{concept is stable}. 
Our scheme is developed and evaluated, both analytically and experimentally, considering the possibility that the data streams are dynamic, i.e., they may experience \emph{concept drift}.

\textbf{ONL: Online learning.}
To deal with dynamic data streams our scheme must learn the aggregation rule "on--the--fly".
In this way the learners maintain an up--to--date aggregation rule and are able to track the concept drifts. 

\textbf{COM: Low complexity.}
Some online ensemble learning schemes, such as \cite{Fan1999, Street2001, Wang2003, 
Masud2009}, need to collect and store chunks of data, that are later processed to update the aggregation model of the system.
This requires a large memory and high computational capabilities, thereby resulting in high implementation cost.
Different from these approaches, in our scheme each data is processed "on--arrival" and afterwards it is thrown away.
Only the up--to--date aggregation model is kept in the memory.
The local prediction of each learner, which is the only information that must be exchanged, consists of a binary value.
Moreover, our scheme is \emph{scalable} to a large number of sources and learners and the learners can be chained in any hierarchical structure. 

\textbf{IND: Independence from local classifiers.}
Different from \cite{Kolter2005, Kolter2007, 
Minku2012}, our scheme is general and can be applied to different types of local classifiers, 
such as support vector machine, decision tree, neural networks, offline/online classifiers, etc.
This feature is important, because the different learners can be managed by different entities, willing to cooperate in exchanging information but not to modify their own local classifiers. 
Also, our algorithm does not need any a priori knowledge about the performance of the local classifiers, it automatically adapts the configuration of the distributed system to the current performance of the local classifiers.

\textbf{DEL: Delayed labels, missing labels, 
and asynchronous learners.}
In distributed environments there are many factors that may impact the performance of the learning system.
First, because obtaining the information about the label may be both costly and time consuming, one cannot expect that all the learners always observe the label in a timely manner. 
Some learners can receive the label with delay, or not receive it at all.
Second, the learners can be asynchronous, i.e., they can observe data at different time instants. 
In this paper we first propose a basic algorithm, considering an idealized scenario in which the above issues are not present, and then we extend our scheme to deal with the above issues.


The rest of this paper is organized as follows.
Section \ref{sec:rel} reviews the existing literature in ensemble learning techniques.
Section \ref{sec:frame} presents our formalism, framework, and algorithm for distributed online learning.
Section \ref{sec:perf} proves a bound for the mis--classification probability of our scheme which depends on the mis--classification probabilities of the best (unknown) static aggregation rule, and of the best (unknown) local classifier. 
Section \ref{sec:dis} discusses several extensions to our learning algorithm to deal with practical issues associated to the distributed implementation of the ensemble of learners, and proves new bounds that apply to these settings.
Section \ref{sec:exp} presents the empirical evaluation of our algorithm on several data sets. 
Section \ref{sec:con} concludes the paper. 

\section{Related Works}
\label{sec:rel}

In this section we review the existing literature on ensemble learning techniques and discuss the differences between the cited works and our paper.

Ensemble learning techniques \cite{Hansen1990, Sewell2008, Alpaydin2010} combine a collection of base classifiers into a unique classifier. 
Adaboost \cite{Freund1997}, for example, trains a sequence of classifiers on increasingly more difficult examples and combines them using a weighted majority rule.
Our paper is clearly different with respect to traditional offline approaches such as Adaboost, which rely on the presence of a training set for offline training the ensemble and assume a stable concept.

An online version of Adaboost is proposed in \cite{Fan1999}.
When a new chunk of data enters the system, the current classifiers are reweighed, a weighted training set is generated, a new classifier (and its weight) is created on this data set, and the oldest classifier is discarded. 
Similar proposals are made in \cite{Street2001, Wang2003, 
Masud2009, Avidan07}.
Our work differs from these online boosting--like techniques because (i) it processes each instance "on arrival" only once, without the need for storage and reprocessing chunks of data, and (ii) it does not require that the local classifiers are centrally retrained (e.g., in a distributed scenario it may be expensive to retrain the local classifiers or unfeasible if the learners are operated by different entities). 

An alternative approach to storing chunks of labeled data consists in updating the ensemble as soon as data flows in the system. 
\cite{Kolter2005} and \cite{Kolter2007} adopt a dynamic weighted majority algorithm, refining, adding, and removing learners based on the global algorithm's performance.
\cite{Minku2012} proposes a scheme based on two online ensembles with different levels of diversity.
The low diversity ensemble is used for system predictions, the high diversity ensemble is used to learn the new concept after a drift is detected.
Our work differs from \cite{Kolter2005, Kolter2007, 
Minku2012} because it does not require that the local classifiers are centrally retrained. 

The literature closest to our work is represented by the multiplicative weight update schemes \cite{Littlestone1994, Littlestone1988, 
Blum1997, Herbster1998} that maintain a collection of given learners, predict using a weighted majority rule, and update online the weights associated to the learners in a multiplicative manner.
Weighted majority \cite{Littlestone1994} decreases the weights of the learners in the pool that disagree with the label whenever the ensemble makes a mistakes.
Winnow2 \cite{Littlestone1988} uses a slightly different update rule, but the final effect is the same as weighted majority.
In \cite{Blum1997} the weights of the learners that agree with the label when the ensemble makes a mistakes are increased, and the weights of the learners that disagree with the label are decreased also when the ensemble predicts correctly.
To prevent the weights of the learners which performed poorly in the past from becoming too small with respect to the other learners, \cite{Herbster1998} proposes a modified version of these schemes adding a phase, after the multiplicative weight update, in which each learner shares a portion of its weight with the other learners.
In our algorithm, differently from \cite{Littlestone1994, Littlestone1988, 
Blum1997, Herbster1998}, the weights are updated in an additive manner and learners can also have negative weights (e.g., a learner that is always wrong would receive a negative weight and could contribute to the system as a learner that is always right).

Finally, we differentiate from all the cited works in another key point: we consider a distributed scenario, allowing for the possibility that the learners observe different data streams. 
This is the reason why in Section \ref{sec:dis} we extend our learning algorithm to address challenges specific to the distributed implementation.

Table \ref{tab:1} summarizes the differences between our approach and the cited works in terms of the features described in Section \ref{sec:intro}. 

\begin{table}
\begin{center}
\caption{Comparison among different ensemble learning works.}
\begin{tabular}{|c|c|c|c|c|c|c|c|c|}
\hline & \textbf{DIS} & \textbf{DYN} & \textbf{ONL} & \textbf{COM} & \textbf{IND} & \textbf{DEL} \\
\hline
\textbf{\cite{Freund1997}} & & & & X & X & \\
\hline
\textbf{\cite{Fan1999, Street2001, Wang2003, Masud2009, Avidan07}} & & X & X & & X & \\
\hline
\textbf{\cite{Littlestone1994, Littlestone1988, Blum1997, Herbster1998}} & & X & X & X & X & \\
\hline
\textbf{\cite{Kolter2005, Kolter2007, Minku2012}} & & X & X & X & & \\
\hline
\textbf{our work} & X & X & X & X & X & X \\
\hline
\end{tabular}
\end{center}
\label{tab:1}
\end{table}

\section{Distributed Learning Framework And The Proposed Algorithm} \label{sec:frame}

We consider a set of $K$ distributed {\em learners}, denoted by $\mathcal{K} = \left\lbrace 1, \ldots, K \right\rbrace $.
Each learner observes a separate sequence of instances.
The time is slotted and the learners are synchronized.
Throughout the paper, we use the indices $i$ and $j$ to denote particular learners, the indices $n$ and $m$ to denote particular time instants, the index $N$ to denote the possible infinite time horizon (i.e., for how many slots the system operates), and bold letters to denote vectors.

At the beginning of each time slot $n$, each learner $i$ observes an instance generated by a \emph{source} $S_i^{(n)}$.
Let $\mathbf{x}_i^{(n)} \in \mathcal{X}_i$ denote the multi--dimensional \emph{instance} observed by learner $i$ at time instant $n$, and $y^{(n)} \in \{ -1 , 1 \}$ denote the corresponding \emph{label}, a common event that the learners have to classify at time instant $n$. 
We call the pair $ ( \mathbf{x}_i^{(n)}, y^{(n)} ) $ a \emph{labeled instance}.
We formally define a source $S_i^{(n)} \triangleq \left\lbrace p_i^{(n)} ( \mathbf{x}_i^{(n)} , y^{(n)}  ) \right\rbrace$ for learner $i$ at time instant $n$ as the probability density function $p_i^{(n)}$ over the labeled instance $( \mathbf{x}_i^{(n)} , y^{(n)}  )$.
We write $\mathbf{S}^{(n)} = \left( S_1^{(n)}, \ldots, S_{K}^{(n)} \right)$ for the \emph{vector of sources} at time instant $n$.

\begin{figure}
     \centering
          \includegraphics[width=\figw]{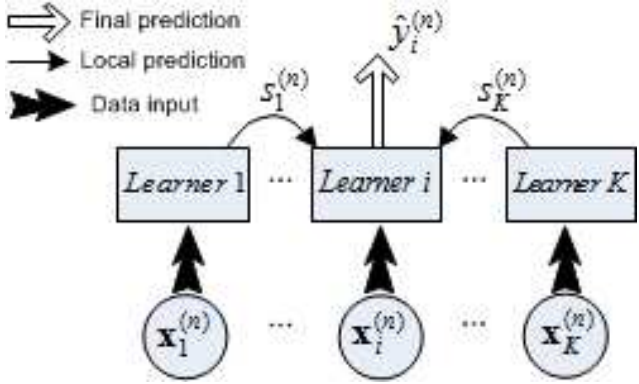}
\caption{System model}
\label{fig:syst}
\end{figure}

The task of a generic learner $i$ at time instant $n$ is to predict the label $y^{(n)}$.
The prediction utilizes the idea of ensemble data mining: each learner adopts an individual classifier to generate a \emph{local prediction}, the local predictions are exchanged, and learner $i$ aggregates its local prediction and the received ones to generate the final prediction $\hat{y}_i^{(n)} \in \{ -1 , 1 \}$.
This process is represented in Fig. \ref{fig:syst}.
Let $s_i^{(n)} \in \{ -1 , 1 \}$ 
denote the local prediction of learner $i$ 
at time instant $n$. 
As in \cite{Littlestone1994, Littlestone1988, 
Blum1997, Herbster1998}, in this paper we assume that the local classifiers are given (i.e., $s_i^{(n)}$ 
is given $\forall \, i \in \mathcal{K}$) and we focus on the adaptivity of the rule that aggregates the local predictions.

Similarly to most ensemble techniques, such as \cite{Freund1997, Littlestone1994, Littlestone1988, 
Blum1997, Fan1999, Wang2003, 
Masud2009, Herbster1998, Kolter2005, Kolter2007}, we consider a \emph{weighted majority} aggregation rule in which learner $i$ maintains a \emph{weight vector} $\mathbf{w}_i^{(n)} \triangleq (w_{i0}^{(n)}, w_{i1}^{(n)}, \ldots, w_{ik}^{(n)}) \in R^{K+1}$, combines it linearly with the \emph{local prediction vector} $\mathbf{s}^{(n)} \triangleq (1, s_1^{(n)}, \ldots, s_k^{(n)})$, and predicts $-1$ if the result is negative, $1$ otherwise, i.e.,
\ba
&\hat{y}_i^{(n)} = \sgn \left( \mathbf{w}_i^{(n)} \cdot \mathbf{s}^{(n)} \right)
= \left\{\begin{array}{ll}
1 & \mbox{if} \; \mathbf{w}_i^{(n)} \cdot \mathbf{s}^{(n)}  \geq 0 \\
-1 & \mbox{otherwise}
\end{array} \right.
\label{eq:dec_rule}
\end{align}
where $\sgn(\cdot)$ is the sign function (we define $\sgn(0) \triangleq 1$) and $\mathbf{w}_i^{(n)} \cdot \mathbf{s}^{(n)}  \triangleq w_{i0}^{(n)} + \sum_{j=1}^K w_{ij}^{(n)} s_j^{(n)}$ is the inner product among the vectors $\mathbf{w}_i^{(n)}$ and $\mathbf{s}^{(n)} $.
The equation $w_{i0}^{(n)} + \sum_{j=1}^K w_{ij}^{(n)} s_j^{(n)} = 0$ defines an hyperplane in $\Re^K$ (the space of the local predictions) which separates the positive predictions (i.e., $\hat{y}_i^{(n)} = 1$) from the negative ones (i.e., $\hat{y}_i^{(n)} = -1$).
Notice that in most of the weighted majority schemes proposed in literature, \cite{Freund1997, Littlestone1994, Littlestone1988, 
Blum1997, Fan1999, Wang2003, 
Masud2009, Herbster1998, Kolter2005, Kolter2007}, $w_{i0}^{(n)} = 0$ which constrains the hyperplane to pass through the origin.
However, in our paper the weight $w_{i0}^{(n)}$ can be thought of as the weight associated to a "virtual learner" that always sends the local prediction $1$, and we introduce it to exploit an additional degree of freedom.

We consider the following rule to update the weight vector $\mathbf{w}_i^{(n)}$ at the end of time instant $n$:
\ba
\mathbf{w}_i^{(n+1)} = \left\{\begin{array}{ll}
\mathbf{w}_i^{(n)} & \mbox{if} \; \hat{y}_i^{(n)} = y^{(n)} \\
\mathbf{w}_i^{(n)} + y^{(n)} \mathbf{s}^{(n)}  & \mbox{otherwise}
\end{array} \right.
\label{eq:percep}
\end{align}
That is, after having observed the true label, learner $i$ compares it with its prediction.
If the prediction is correct, the model is not modified.
If the prediction is incorrect, the weights of the learners that reported a wrong prediction are decreased by one unit, whereas the weights of the learners that reported a correct prediction are increased by one unit.\footnote{This is in the same philosophy of many weighted majority schemes \cite{Littlestone1994, Littlestone1988, 
Herbster1998} and boosting--like techniques \cite{Fan1999, Street2001, Wang2003, 
Masud2009} that improve the model focusing mainly on those instances in which the actual model fails.} 

Since (\ref{eq:percep}) is analogous to the learning rule of a Perceptron algorithm \cite{Rosenblatt1957}, we call the resulting online learning scheme Perceptron Weighted Majority (PWM). 
We initialize to $0$ the weights $w_{ij}^{(1)}$, $i, j \in \mathcal{K}$.
Because at the end of each time instant $n$ the value of $w_{ij}^{(n)}$ can remain constant, decrease by one unit, or increase by one unit, $w_{ij}^{(n)}$ is always an integer number.


\begin{algorithm}
\caption{Perceptron Weighted Majority (PWM)} 
\begin{algorithmic}[1]
\STATE \textbf{Initialization}: $w_{ij} = 0$, $\forall \, i,j \in \mathcal{K}$
\STATE  \textbf{For} each learner $i$ and time instant $n$ 
\STATE ~~Observe $\mathbf{x}_i^{(n)}$
\STATE ~~Obtain $\mathbf{s}^{(n)} = (1, s_1^{(n)}, \ldots, s_k^{(n)})$
\STATE ~~Predict $\hat{y}_i^{(n)} \leftarrow \sgn(\mathbf{w}_i \cdot \mathbf{s}^{(n)})$
\STATE ~~Observe $y^{(n)}$ 
\STATE ~~\textbf{If} $y ^{(n)}\neq \hat{y}_i^{(n)}$ \textbf{do} $\mathbf{w}_i \leftarrow \mathbf{w}_i + y^{(n)} \mathbf{s}^{(n)}$
\end{algorithmic}
\label{pwm}
\end{algorithm}

\begin{figure}
     \centering
          \includegraphics[width=\figw]{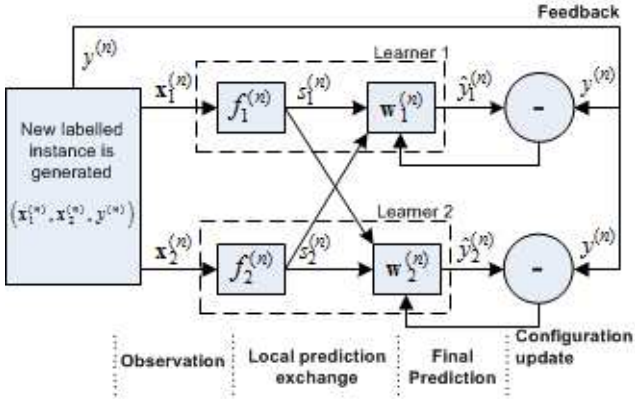}
\caption{Illustrative system of two learners adopting the PWM scheme}
\label{fig:syst2}
\end{figure}

To summarize, the sequence of events that take place at time instant $n$ for each learner $i$ adopting the PWM algorithm can be described as follows. 
\begin{itemize}
\item[\textbf{1.}] \textbf{Observation:} learner $i$ observes the instance $\mathbf{x}_i^{(n)}$;
\item[\textbf{2.}] \textbf{Local prediction exchange:} learner $i$ sends its local prediction $s_i^{(n)} = f_i^{(n)} (\mathbf{x}_i^{(n)})$ to the other learners, and receives the local predictions $\mathbf{s}_{j}^{(n)} = f_j^{(n)} (\mathbf{x}_j^{(n)})$, $\forall \, j \neq i$, from the other learners; 
\item[\textbf{3.}] \textbf{Final prediction:} learner $i$ computes and outputs its final prediction $\hat{y}_i^{(n)} = \sgn \left( \mathbf{w}_i^{(n)} \cdot \mathbf{s}^{(n)} \right)$;
\item[\textbf{4.}] \textbf{Feedback:} learner $i$ observes the true label $y^{(n)}$;
\item[\textbf{5.}] \textbf{Configuration update:} learner $i$ updates the weight vector $\mathbf{w}_i^{(n)}$ adopting (\ref{eq:percep}).
\end{itemize}
Fig. \ref{fig:syst2} illustrates this sequence of events for a system of two learners.

\section{Performance of PWM} \label{sec:perf}

In this section we analytically quantify the performance of PWM in terms of its \emph{empirical mis--classification probability} (shortly, mis--prediction probability), which is defined as the number of prediction mistakes per instance. 

We prove two upper bounds for the mis--classification probability of our scheme. 
The first bound depends on the mis--classification probability of the best (unknown) static aggregation rule, and is particularly useful when the local classifiers are \emph{weak} (i.e., their performance are comparable to random guessing) but their combination can result in an accurate ensemble.\footnote{It is known that the combination of weak classifiers can result in a high accurate ensemble \cite{Opitz99}, in particular when the classifiers are diverse and their errors are independent.} 
The second bound depends on the mis--classification probability of the best (unknown) local classifiers, and is particularly useful when there are accurate local classifiers in the system.
We then combine these two bounds into a unique bound.
We show that the resulting bound and the mis--classification probability of PWM tend asymptotically to $0$ if the mis--classification probability of the best static aggregation rule or the mis--classification probability of the best local classifier tend to $0$.
Then we formally define the notions of \emph{concept} and \emph{concept drift} and we show that the mis--classification probability of PWM tends to $0$ if, for each concept, there exists a (unknown) static aggregation rule whose mis--classification probability (for the considered concept) tends to $0$.

Importantly, we remark that PWM is designed in absence of a priori knowledge about the sources and the performance of the local classifiers.
We do not need to know a priori whether there are accurate local classifiers or accurate aggregation rules.
It is the scheme itself that adapts the configuration of the distributed system to the current performance of the local classifiers.

\subsection{Definitions} \label{sec:def}

Given the sequence of $N$ labeled instances 
$$
\mathbf{D}_N \triangleq \left( \mathbf{x}_1^{(n)}, \ldots , \mathbf{x}_{K}^{(n)}, y^{(n)} \right)_{n=1, \ldots, N} \;\; ,
$$
we denote by $P_{i}(\mathbf{D}_N)$ the mis--classification probability of the local classifier used by learner $i$, by $P^*(\mathbf{D}_N)$ the mis--classification probability of the most accurate local classifier, and by $v^*(\mathbf{D}_N)$ the number of local classifiers whose mis--classification probabilities are $P^*(\mathbf{D}_N)$,\footnote{This paper does not distinguish among different classification errors, i.e., among false alarms and mis--detections.}
\ba
P_{i}(\mathbf{D}_N) &\triangleq \dfrac{1}{N} \sum_{n=1}^{N} I \{ s_i \neq y^{(n)} \}  \nonumber\\
P^*(\mathbf{D}_N) &\triangleq \min_{i \in \mathcal{K}} P_{i}(\mathbf{D}_N) \nonumber\\
v^*(\mathbf{D}_N) &\triangleq \abs{ \left\lbrace i: P_{i}(\mathbf{D}_N)=P^*(\mathbf{D}_N) \right\rbrace } \nonumber
\end{align}
where $\abs{\cdot}$ denotes the cardinality of the considered set.

Also, we denote by $P^O(\mathbf{D}_N)$ the mis--classification probability of learner $i$ if it combines the local predictions of all the learners using the \emph{optimal static weight vector} $\mathbf{w}^O$ that minimizes its number of mistakes,
\ba
P^O(\mathbf{D}_N) &\triangleq \min_{\mathbf{w}^O} \dfrac{1}{N} \sum_{n=1}^{N} I \{ \sgn \left( \mathbf{w}^O \cdot \mathbf{s}^{(n)} \right) \neq y^{(n)} \} \nonumber
\end{align}

\begin{rem}
$P^O(\mathbf{D}_N)$ and $\mathbf{w}^O$ are the same for all the learners.
For this reason, we do not use the subscript $i$. 
\end{rem}

\begin{rem}
$P^O(\mathbf{D}_N) \leq P^*(\mathbf{D}_N)$, in fact it is always possible to select a static weight vector such that the final prediction in each time instant $n$ is equal to the prediction of the best classifier.
\end{rem}

\begin{rem}
The computation and adoption of $\mathbf{w}^O$ would require to know in advance, at the beginning of time instant $1$, the sequences of local predictions $\mathbf{s}^{(n)}$ and labels $y^{(n)}$, for every time instant $n = 1 , \ldots, N$.
\end{rem}

Moreover, we denote by $P_{i}^{PWM}(\mathbf{D}_N)$ the mis--classification probability of learner $i$ if it adopts the $PWM$ scheme, 
\ba
P_{i}^{PWM}(\mathbf{D}_N) &\triangleq \dfrac{1}{N} \sum_{n=1}^{N} I \{ \sgn \left( \mathbf{w}^{(n)} \cdot \mathbf{s}^{(n)} \right) \neq y^{(n)} \} \nonumber
\end{align}
where $w_{ij}^{(1)}=1$, $\forall \, i, j$, and $\mathbf{w}_i^{(n)}$ evolves according to (\ref{eq:percep}).
We denote by $P^{PWM}(\mathbf{D}_N)$ the average mis--classification probability of the distributed system if all the learners adopt the $PWM$ scheme,
\ba
P^{PWM}(\mathbf{D}_N) &\triangleq \dfrac{1}{K} \sum_{i=1}^{K} P_{i}^{PWM}(\mathbf{D}_N) 
\label{eq:P_sys}
\end{align}

\begin{rem}
In this section $P^{PWM}(\mathbf{D}_N) = P_i^{PWM}(\mathbf{D}_N)$, $\forall \, i$, because the weight vectors of the learners are equally initialized and we assumed that the learners are synchronized and always observe the labels, hence $\mathbf{w}_i^{(n)}$ and $\mathbf{w}_j^{(n)}$ evolve in the same way and $P_i^{PWM}(\mathbf{D}_N) = P_j^{PWM}(\mathbf{D}_N)$, $\forall \, i,j$.
However, in Section \ref{sec:dis} we describe several extensions to our online learning algorithm, in which $\mathbf{w}_i^{(n)}$ and $\mathbf{w}_j^{(n)}$ evolve differently, and consequently $P_i^{PWM}(\mathbf{D}_N) \neq P_j^{PWM}(\mathbf{D}_N)$, $i \neq j$.
\end{rem}


\subsection{Bounds for PWM mis--classification probability} \label{sec:def}

In this subsection we derive the following results.
Lemma \ref{lem:1} proves a bound for $P^{PWM}(\mathbf{D}_N)$ as a function of $P^O(\mathbf{D}_N)$.
Lemma \ref{lem:2} proves a bound for $P^{PWM}(\mathbf{D}_N)$ as a function of $P^*(\mathbf{D}_N)$.
Theorem \ref{teo:1} combines these two bounds into a unique bound.
Finally, as a special case of Theorem \ref{teo:1}, Theorem \ref{teo:2} shows that $P^{PWM}(\mathbf{D}_N)$ converges to $0$ if $P^*(\mathbf{D}_N)$ or $P^O(\mathbf{D}_N)$ converge to $0$.

\begin{lemma}
For every sequence of labeled instances $\mathbf{D}_N$, the mis--classification probability $P^{PWM}(\mathbf{D}_N)$ is bounded by 
\be
\mathbf{B}_1(\mathbf{D}_N) \triangleq 2 K  P^O(\mathbf{D}_N) + \dfrac{K (K + 1)}{N} \nonumber
\ee
\label{lem:1}
\end{lemma}
\begin{IEEEproof}
See Appendix \ref{app:1}.
\end{IEEEproof}

\begin{rem}
Lemma \ref{lem:1} shows that it is not always beneficial to have many learners in the system. 
On one hand, an additional learner can decrease the benchmark prediction probability $P^O(\mathbf{D}_N)$.
On the other hand, it increases the number of learners $K$, and as a consequence the maximum number or errors needed to approach the benchmark weight vector $\mathbf{w}^O$ increases.
The final impact on $P^{PWM}(\mathbf{D}_N)$ depends on which of the two effects is the strongest.
\end{rem}

\begin{rem}
If the optimal static weight vector $\mathbf{w}^O$ allows to predict always correctly the labeled instances $\mathbf{D}_N$, i.e., $P^O(\mathbf{D}_N)=0$, then $P^{PWM}(\mathbf{D}_N) \leq \frac{K (K + 1)}{N}$. Hence, the bound increases quadratically in the number of learners $K$, but decreases linearly in the number of instances $N$.
\label{rem:1}
\end{rem}

We define the function
\be
f(x, y) \triangleq 2 x + \frac{K + 1}{2 N y} +\sqrt{\left(\frac{K + 1}{2 N y}\right)^2 + \frac{2 (K + 1) x}{N y}} \nonumber
\ee

\begin{lemma}
For every sequence of labeled instances $\mathbf{D}_N$, the mis--classification probability $P^{PWM}(\mathbf{D}_N)$ is bounded by 
\be
\mathbf{B}_2(\mathbf{D}_N) \triangleq f\left( P^*(\mathbf{D}_N), v^*(\mathbf{D}_N) \right) \nonumber
\ee
\label{lem:2}
\end{lemma}

\begin{IEEEproof}
See Appendix \ref{app:2}.
\end{IEEEproof}

\begin{rem}
If the best local classifier always predicts correctly the labeled instances $\mathbf{D}_N$, i.e., $P^*(\mathbf{D}_N)=0$, then $P^{PWM}(\mathbf{D}_N) \leq \frac{K + 1}{N \cdot v^*(\mathbf{D}_N)}$. 
This bound is $K \cdot v^*(\mathbf{D}_N)$ times better than the bound in Remark \ref{rem:1}.
\end{rem}

\begin{rem}
Asymptotically, for $N \to + \infty$, $\mathbf{B}_1(\mathbf{D}_N) \to 2 K  P^O(\mathbf{D}_N)$ and  $\mathbf{B}_2(\mathbf{D}_N) \to 2 P^*(\mathbf{D}_N)$.
On one hand, if the local classifiers are weak (i.e., $P^*(\mathbf{D}_N) \simeq 0.5$) but their aggregation is very accurate (i.e., $P^O(\mathbf{D}_N) \ll 1$), the first bound is usually stricter than the second.
On the other hand, if the performance of the best local classifier is comparable with the performance of the optimal static aggregation rule (i.e., $P^*(\mathbf{D}_N) \simeq P^O(\mathbf{D}_N)$), the second bound is $K$ times stricter than the first one.
Notice that also the bound computed in \cite{Littlestone1994}, for the multiplicative update rule, depends linearly on the accuracy of the best classifier.
\label{rem:8}
\end{rem}

In the following theorem, we combine $\mathbf{B}_1(\mathbf{D}_N)$ and $\mathbf{B}_2(\mathbf{D}_N)$ into a unique bound.

\begin{theorem}
For every sequence of labeled instances $\mathbf{D}_N$, the mis--classification probability $P^{PWM}(\mathbf{D}_N)$ is bounded by 
\be
\mathbf{B}(\mathbf{D}_N) \triangleq \min \left\lbrace \mathbf{B}_1(\mathbf{D}_N)  , \mathbf{B}_2(\mathbf{D}_N) , 1 \right\rbrace \nonumber
\ee
\label{teo:1}
\end{theorem}

\begin{IEEEproof}
We simply combine Lemmas \ref{lem:1} and \ref{lem:2}, and the fact that the mis--prediction probability cannot be larger than $1$.
\end{IEEEproof}

Importantly, notice that the bound $\mathbf{B}(\mathbf{D}_N)$ is valid for any time horizon $N$ and for any sequence of labeled instances $\mathbf{D}_N$.  
As a particular case, if the time horizon tends to infinity and there exists either 1) a static aggregation weight vector whose mis--classification probability tends to $0$ (i.e., $P^O(\mathbf{D}_N) \to 0$), or 2) a local classifier whose mis--classification probability tends to $0$ (i.e., $P^*(\mathbf{D}_N) \to 0$), we obtain that the mis--classification probability of PWM tends to $0$ as well.
Notice that $P^*(\mathbf{D}_N) \to 0$ is a specific case of $P^O(\mathbf{D}_N) \to 0$, because $P^O(\mathbf{D}_N) \leq P^*(\mathbf{D}_N)$.
Hence, in the statement of the following theorem we consider only the case $P^O(\mathbf{D}_N) \to 0$.

\begin{theorem}
If 
$\lim_{N \to + \infty} P^O(\mathbf{D}_N) = 0$,
then 
\be
\lim_{N \to + \infty} P^{PWM}(\mathbf{D}_N) = 0 \nonumber
\ee
\label{teo:2}
\end{theorem}

\begin{IEEEproof}
$P^{PWM}(\mathbf{D}_N) \leq 2 K  P^O(\mathbf{D}_N) + \frac{K (K + 1)}{N}$ and the right hand side tends to $0$ for $N \to + \infty$.
\end{IEEEproof}

\subsection{Bound in the Presence of Concept Drifts} \label{sec:drift}

Given two time instants $n$ and $m$, $n > m$, we write $S_i^{(n)} = S_{i}^{(m)}$ if the labeled instances $ ( \mathbf{x}_i^{(n)}, y_i^{(n)} )$ and $( \mathbf{x}_i^{(m)}, y_i^{(m)} ) $ are independently sampled from the same distribution.
We write $\mathbf{S}^{(n)} = \mathbf{S}^{(m)}$ if $S_i^{(n)} = S_i^{(m)}$, $\forall \, i \in \mathcal{K}$.
As in \cite{Zliobaite10}, we refer to a particular vector of sources as a \emph{concept}.
The expression \emph{concept drift} \cite{Zliobaite10, Babcock2002, Gao2007, Street2001, Wang2003, 
Masud2009, Herbster1998, Kolter2005, Kolter2007, 
Minku2012, Ho10, Crammer10} refers to a change of concept that occurs in a certain time instant. 
According to \cite{Zliobaite10}, we say that at time instant $n$ there is a concept drift if $\mathbf{S}^{(n+1)} \neq \mathbf{S}^{(n)}$.

Theorem \ref{teo:2} states that $P^{PWM}(\mathbf{D}_N) \to 0$ if $P^O(\mathbf{D}_N) \to 0$.
Unfortunately, in presence of concept drifts it is highly improbable that $P^O(\mathbf{D}_N) \to 0$. 
In fact, the accuracies of the local classifiers can change consistently from one concept to another, and the best weight vector to aggregate the local predictions changes accordingly. 
In the following we generalize the result of Theorem \ref{teo:2} considering an assumption that is more realistic if there are concept drifts.

We denote by $\mathbf{D}_{N_c}$ a sequence of $N_c$ labeled instances generated by the concept $\mathbf{S}_c^{(n)}$.
We say that the concept $\mathbf{S}_c^{(n)}$ is \emph{learnable} if, $\forall \, \mathbf{D}_{N_c}$,
\ba
\lim_{N_c \to + \infty} \min_{\mathbf{w}_{i,c}^O} \dfrac{1}{N_c} \sum_{n=1}^{N_c} I \{ \sgn \left( \mathbf{w}_{i,c}^O \cdot \mathbf{s}^{(n)} \right) \neq y^{(n)} \} = 0  \nonumber
\end{align}
That is, the concept $\mathbf{S}_c^{(n)}$ is learnable if there exists a static weight vector $\mathbf{w}_{i,c}^O$ whose asymptotic mis--classification probability, over the labelled instances generated by that concept, tends to $0$. 

\begin{theorem}
If $\mathbf{D}_{N}$, for $N \to + \infty$, is generated by a finite number of learnable concepts and a finite number of concept drifts occurred, then
\be
\lim_{N \to + \infty} P^{PWM}(\mathbf{D}_N) \to 0 \nonumber
\ee
\label{teo:3}
\end{theorem}

\begin{IEEEproof}
See Appendix \ref{app:3}.
\end{IEEEproof}

\begin{rem}
Theorem \ref{teo:2} requires the existence of a \emph{unique weight vector}, $\mathbf{w}^O$, whose mis--classification probability over the labeled instances generated by \emph{all} concepts converges to $0$.
Theorem \ref{teo:3} requires the existence of \emph{one weight vector for concept}, $\mathbf{w}_{i,c}^O$, whose mis--classification probability over the labeled instances generated by concept $\mathbf{S}_c^{(n)}$ converges to $0$.
\end{rem}

\section{Extended PWM} \label{sec:dis}

So far we have considered an idealized setting in which all the learners always observe an instance at the beginning of the time instant (i.e., they are synchronous), and they always observe the corresponding label at the end of the time instant. 
In a distributed environment one cannot expect that these assumptions are always satisfied: sometimes the learners can be asynchronous, receive the label with delay, or not receive it at all.
In this section we address these challenges proposing, for each of them, a modification to the basic PWM scheme introduced in Section \ref{sec:frame},  and we extend Theorems \ref{teo:1} and \ref{teo:3} for each modified version of PWM.\footnote{Notice that Theorem \ref{teo:3} is a more general version of Theorem \ref{teo:2}, and hence we do not need to extend also Theorem \ref{teo:2}.} 
At the end of this section we explicitly write the \emph{extended PWM} algorithm that includes all the proposed modification to jointly deal with all the considered challenges. 


\subsection{Delayed and Out--Of--Order Labels}  \label{sec:deay}

In some cases the true label corresponding to a time instant $n$ is observed with delay.
For example, in a distributed environment one learner can observe the label immediately, and communicate it to the other learners at a later stage.
In this subsection we show that our algorithm can be modified in order to deal with this situation, with a price to pay in terms of increased memory.

We denote by $d_i^{(n)}$ the number of time slots after which learner $i$ observes the $n$--th label. 
We assume that $d_i^{(n)}$ is not known a priori, but is bounded by a maximum delay $\overline{d}_i$, $\forall \, n$, which is known.
Also, we allow for the possibility that the labels are received out of order (e.g., it is possible that learner $i$ observes the label $y^{(n+1)}$ before the label $y^{(n)}$), but we assume that, when a label is received, the time instant it refers to is known.

PWM is modified as follow.
Learner $i$ maintains in memory all the local prediction vectors that refer to the not yet observed labels.
As soon as learner $i$ receives the label $y^{(m)}$, it computes the prediction $\hat{y}_i^{(m)} = \sgn ( \mathbf{w}_i^{(n)} \cdot \mathbf{s}^{(m)} )$ which it would have made at time instant $m$ with the current weight vector $\mathbf{w}_i^{(n)}$, and updates the weight vector according to 
\ba
\mathbf{w}_i^{(n+1)} = \left\{\begin{array}{ll}
\mathbf{w}_i^{(n)} & \mbox{if} \; \hat{y}_i^{(m)} = y^{(m)} \\
\mathbf{w}_i^{(n)} + y^{(m)} \mathbf{s}^{(m)}  & \mbox{otherwise}
\end{array} \right. \nonumber
\end{align}
This update rule is similar to (\ref{eq:percep}), but now the updates may happen with delays.
In particular, since different learners experience different delays, the weight vectors $\mathbf{w}_i^{(n)}$ and $\mathbf{w}_j^{(n)}$, $i \neq j$, follow different dynamics.


\begin{theorem}
For every sequence of labeled instances $\mathbf{D}_N$, $P^{PWM}(\mathbf{D}_N)$ is bounded by
\be
\mathbf{B}(\mathbf{D}_N) + \dfrac{\sum_{i=1}^K \overline{d}_i}{N K} \nonumber
\ee
\label{teo:4}
\end{theorem}

\begin{IEEEproof}
See Appendix \ref{app:4}.
\end{IEEEproof}

\begin{rem}
The term $\frac{\sum_{i=1}^K \overline{d}_i}{N K}$ can be interpreted as the maximum loss for the delayed labels.
\end{rem}

\begin{theorem}
If $\mathbf{D}_{N}$, for $N \to + \infty$, is generated by a finite number of learnable concepts and a finite number of concept drifts occurred, then
\be
\lim_{N \to + \infty} P^{PWM}(\mathbf{D}_N) \to 0 \nonumber
\ee
\label{teo:5}
\end{theorem}

\begin{IEEEproof}
See Appendix \ref{app:4bis}.
\end{IEEEproof}

\subsection{Missing Labels}  \label{sec:part_obs}

In a distributed environment one cannot expect that all the learners always receive the label, in particular in those scenarios in which obtaining the information about the label may be both costly and time consuming. 
In this subsection we show that our scheme can be easily extended to deal with situations in which the true labels are only occasionally observed.

Let $g_i^{(n)} \triangleq 1$ if learner $i$ observes the label $y^{(n)}$ at the end to time instant $n$, $g_i^{(n)} \triangleq 0$ otherwise.
The following update rule represents the natural extension of (\ref{eq:percep}) to deal with missing labels: 
\ba
\mathbf{w}_i^{(n+1)} = \left\{\begin{array}{ll}
\mathbf{w}_i^{(n)} & \mbox{if} \; g_i^{(n)} = 0 \; \mbox{or} \; \hat{y}_i^{(n)} = y^{(n)} \\
\mathbf{w}_i^{(n)} + y^{(n)} \mathbf{s}^{(n)}  & \mbox{otherwise}
\end{array} \right. \nonumber
\end{align}
That is, learner $i$ updates the weight vector $\mathbf{w}_i^{(n)}$ only when it observes the true label and it recognizes it made a prediction error.
Notice that different learners observe different labels; therefore, the weight vectors $\mathbf{w}_i^{(n)}$ and $\mathbf{w}_j^{(n)}$, $i \neq j$, follow different dynamics.

Now we consider a simple model of missing labels and we derive the equivalent for the Theorems \ref{teo:1} and \ref{teo:3}. 
We assume that $g_i^{(n)}$ is an independent and identically distributed (i.i.d.) process, $\forall \, i$, and denote by $\mu$ the probability that $g_i^{(n)}=1$, $0 < \mu < 1$.\footnote{We can extend the analysis considering an observation probability $\mu_i$ that depends on the learner $i$. The results would be similar to those obtained with a unique $\mu$, but the notations would be much messier.}
That is, at the end of a generic time instant $n$ learner $i$ observes the label with probability $\mu$.

Denote by $N_e^{PWM}$ the number of prediction errors observed by learner $i$, i.e., the number of times $i$ observes the label and recognizes it made a prediction mistake.
We define the function
\be
\lambda \left( y , z\right) \triangleq \sqrt{\dfrac{1}{2 z} \ln \dfrac{1}{y} }
\label{eq:lambda}
\ee

\begin{theorem}
Given the sequence of instances $\mathbf{D}_N$, for any level of confidence $\epsilon > 0$ such that $\lambda \left( \epsilon , \tilde{L}_i^{PWM} \right) \leq \mu$, with probability at least $1 - \epsilon$ we have that $P^{PWM}(\mathbf{D}_N)$ is bounded by
\be
\dfrac{\mathbf{B}(\mathbf{D}_N) }{\mu - \lambda \left( \epsilon , N_e^{PWM} \right)}  
\label{eq:boundR5}
\ee
\label{teo:6}
\end{theorem}

\begin{IEEEproof}
See Appendix \ref{app:5}.
\end{IEEEproof}

\begin{rem}
The denominator $\mu - \lambda ( \epsilon , N_e^{PWM} )$, which is lower than $1$, can be interpreted as the maximum loss for the missing labels.
Notice that, for any given level of confidence $\epsilon$, the function $\lambda \left( \epsilon , N_e^{PWM} \right)$ is decreasing in the number of observed errors $N_e^{PWM}$, and tends to $0$ if $N_e^{PWM} \to + \infty$. As a consequence, the bound (\ref{eq:boundR5}) tends to $\mathbf{B}(\mathbf{D}_N)$ divided by the probability to observe a label $\mu$.
\end{rem}

\begin{theorem}
If $\mathbf{D}_{N}$, for $N \to + \infty$, is generated by a finite number of learnable concepts and a finite number of concept drifts occurred, then
\be
\lim_{N \to + \infty} P^{PWM}(\mathbf{D}_N) \to 0
\label{eq:asympt3}
\ee
\label{teo:7}
\end{theorem}

\begin{IEEEproof}
See Appendix \ref{app:5bis}.
\end{IEEEproof}

\subsection{Asynchronous Learners}  \label{sec:async}

Another important factor that may impact the performance of an online learning distributed system is the synchronization among the learners. 
So far we have assumed that each learner observes an instance in every time instant. 
However, in many practical scenarios different learners may capture instances in different time instants, and they can have different acquisition rates. 
In this subsection we extend our scheme to deal with this situation. 

PWM is modified as follow. 
A learner does not send a local prediction when it does not observe the instance; however, it can still output a final prediction exploiting the local predictions received from the other learners. 
A generic learner $i$ maintains two weight vectors: $\mathbf{w}_{i,s}^{(n)}$ and $\mathbf{w}_{i,a}^{(n)}$. 
At the time instants in which all the learners observe the instances (i.e., when the learners are synchronized), learner $i$ aggregates all the local predictions using $\mathbf{w}_{i,s}^{(n)}$ and then, after having observed the label, updates $\mathbf{w}_{i,s}^{(n)}$ using (\ref{eq:percep}).
At the time instants in which some learners do not observe the instances (i.e., when the learners are not synchronized), learner $i$ set to $0$ the non received local predictions (i.e., it treats the learners that do not observe the instances as "abstainer"), aggregate the local predictions using $\mathbf{w}_{i,a}^{(n)}$, and then, after having observed the label, updates $\mathbf{w}_{i,a}^{(n)}$ using (\ref{eq:percep}) (notice that the weights of the abstainers are not modified).

Given the sequence of labelled instances $\mathbf{D}_{N}$, we denote by $M$ the number of times in which the instances are jointly observed by all the learners.
We define the synchronization index $\alpha \triangleq \frac{N - M}{N}$.
Notice that $0 \leq \alpha \leq 1$, the lower  $\alpha$ the more synchronized the learners.

\begin{theorem}
Given the sequence of instances $\mathbf{D}_N$, $P^{PWM}(\mathbf{D}_N)$ is bounded by
\be
\mathbf{B}(\mathbf{D}_N)  + \alpha \nonumber
\ee
\label{teo:10}
\end{theorem}

\begin{IEEEproof}
See Appendix \ref{app:7}.
\end{IEEEproof}

\begin{rem}
The synchronization index $\alpha$ can be interpreted as the maximum loss for non synchronized learners.
If the learners are always synchronous (i.e., $\alpha = 0$), Theorem \ref{teo:10} is equal to Theorem \ref{teo:1}.
\end{rem}

\begin{theorem}
If $\mathbf{D}_{N}$, for $N \to + \infty$, is generated by a finite number of learnable concepts and a finite number of concept drifts occurred, then
\be
\lim_{N \to + \infty} P^{PWM}(\mathbf{D}_N) \leq \alpha
\label{eq:asympt5}
\ee
\label{teo:11}
\end{theorem}

\begin{IEEEproof}
See Appendix \ref{app:7bis}.
\end{IEEEproof}

\begin{rem}
Different from Theorems \ref{teo:3}, \ref{teo:5}, and \ref{teo:7}, in Theorem \ref{teo:11} the mis--classification probability does not tend to $0$. 
In fact, the consequence of non synchronized learners is that a learner does not have, in all the time instances, the local predictions of all the other learners, and this lack of information may result in a mis--classification. 
\end{rem}

\begin{rem}
Theorem \ref{teo:11} can be used as a tool to design the acquisition protocol adopted by the learners.
If we know that the concepts are learnable and we have to satisfy a mis--classification probability constraint $P_{mis}$, Eq. (\ref{eq:asympt5}) can be used to choose the acquisition protocol such that the synchronization index $\alpha$ is equal to or lower than $P_{mis}$.
\end{rem}

\begin{algorithm}
\caption{Extended PWM} 
\begin{algorithmic}
\STATE \textbf{Initialization}: $w_{ij,s} = w_{ij,a} = 0$, $\forall \, i,j \in \mathcal{K}$
\STATE  \textbf{For} each learner $i$ and time instant $n$ 
\STATE ~~\textbf{If} $s_j^{(n)}$ is received $\forall \, j \in \mathcal{K}$
\STATE ~~~~$\hat{y}_i^{(n)} \leftarrow \sgn(\mathbf{w}_{i,s} \cdot \mathbf{s}^{(n)})$
\STATE ~~\textbf{Else} 
\STATE ~~~~\textbf{For} each $j$ such that $s_j^{(n)}$ is not received \textbf{do} $s_j^{(n)} \leftarrow 0$ 
\STATE ~~~~$\hat{y}_i^{(n)} \leftarrow \sgn(\mathbf{w}_{i,a} \cdot \mathbf{s}^{(n)})$
\STATE ~~\textbf{For} each instant $m \leq n$ such that $y^{(m)}$ is observed
\STATE ~~~~\textbf{If} $s_j^{(m)} \neq 0$ $\forall \, j$ 
\STATE ~~~~~~\textbf{If} $y^{(m)} \neq \sgn(\mathbf{w}_{i,s} \cdot \mathbf{s}^{(m)})$ \textbf{do} $\mathbf{w}_{i,s} \leftarrow \mathbf{w}_{i,s} + y^{(m)} \mathbf{s}^{(m)}$
\STATE ~~~~\textbf{Else}
\STATE ~~~~~~\textbf{If} $y^{(m)} \neq \sgn(\mathbf{w}_{i,a} \cdot \mathbf{s}^{(m)})$ \textbf{do} $\mathbf{w}_{i,a} \leftarrow \mathbf{w}_{i,a} + y^{(m)} \mathbf{s}^{(m)}$
\label{pwm}
\end{algorithmic}
\label{pwm}
\end{algorithm}

\section{Experiments} \label{sec:exp}

In this section we evaluate empirically the basic PWM algorithm and the extended PWM algorithm we proposed in Sections \ref{sec:frame} and \ref{sec:dis}, respectively. 
In order to compare PWM with other state--of--the--art ensemble learning techniques that do not deal with a distributed environment, in the first set of experiments (Subsection \ref{sec:centr}) all the learners observe the same data stream, but they are pre--trained on different data sets and hence their local predictions are in general different. 
In the second set of experiments (Subsection \ref{sec:distr}), different learners observe different data streams.
In this case we compare PWM against a learner that predicts using only its local prediction, and analyze the impact on their performance of delayed labels, missing labels, and asynchronous learners.

\subsection{Unique Data Stream}
\label{sec:centr}

In this subsection we test PWM and other state--of--the--art solutions using real data sets that are generated from a unique data stream. 
First, we shortly describe the data sets, then we discuss the results.

\subsubsection{Real Data Sets}
\label{sec:sets}

We consider four data sets, well known in the data mining community, that refer to real--world problems.
In particular, the first three data sets are widely used by the literature dealing with concept drift (which is the closest to our work), because they exhibit evident drifts.

\textbf{R1: Network Intrusion.} The network intrusion data set, used for the KDD Cup $1999$ and available in the UCI archive \cite{UCI}, consists of a series of TCP connection records, labeled either as normal connections or as attacks. For a more detailed description of the data set we refer the reader to \cite{Gao2007}, that shows that the network intrusion data set contains non-stationary data.
This data set is widely used in the stream mining literature dealing with concept drift \cite{Gao2007, Masud2009, Minku2012, Wu12}.

\textbf{R2: Electricity Pricing.} The electricity pricing data set holds information for the Australian New South Wales electricity market. The binary label (up or down) identifies the change of the price relative to a moving average of the last 24 hours. For a more detailed description of this dataset we refer the reader to \cite{Harries99}. An appealing property of this data set is that it contains drifts of different types, due to changes in consumption habits, the seasonability, and the expansion of the electricity market. This data set is widely used in the stream mining literature dealing with concept drift \cite{Harries99, Gama04, Baena06, Kuncheva08, Bifet10, Kolter2007, Minku2012, Bifet07}. 

\textbf{R3: Forest Cover Type.} The forest cover type data set from UCI archive \cite{UCI} contains cartographic variables of four wilderness areas of the Roosevelt National Forest in northern Colorado. Each instance refers to a $30 \times 30$ meter cell of one of these areas and is classified with one of seven possible classes of forest cover type. Our task is to predict if an instance belong to the first class or to the other classes. For a more detailed description of this dataset we refer the reader to \cite{Blackard98}. The forest cover type data set contains drifts because data are collected in four different areas. This data set is widely used in the stream mining literature dealing with concept drift \cite{Masud2009, Bifet10, Gama03, Oza01}.

\textbf{R4: Credit Card Risk Assessment.} In the credit card risk assessment data set, used for the PAKDD $2009$ Data Mining Competition \cite{PAKDD09}, each instance contains information about a client that accesses to credit for purchasing on a specific retail chain. The client is labeled as good if he was able to return the credit in time, as bad otherwise. For a more detailed description of this dataset we refer the reader to \cite{PAKDD09}. This data set does not contain drifts because the data were collected during one year with a stable inflation condition. In fact, to the best of our knowledge, the only work dealing with concept drift that uses this data set is \cite{Minku2012}.


\subsubsection{Results}
\label{sec:res2}

In this experiment we compare our scheme with other state--of--the--art ensemble learning algorithms.
Table \ref{tab:schemes} lists the considered algorithms, the corresponding references, the parameters we adopted (that are equal to the ones used in the corresponding papers, except for the window size that is obtained following a tuning procedure), and their performance in the considered data sets.
We shortly described these algorithms in Section \ref{sec:rel}, for a more detailed description we refer the reader to the cited literature.

\begin{table*}
\caption{The considered schemes, their parameters, and their percentages of mis--classifications in the data sets \textbf{R1}--\textbf{R4} }
\begin{center}
\begin{tabular}{|c|c|c|c|c|c|c|c|c|c|c|c|}
\hline
\multirow{2}{*}{\textbf{Abbreviation}} & \multirow{2}{*}{\textbf{Name of the Scheme}} & \multirow{2}{*}{\textbf{Reference}} & \multirow{2}{*}{\textbf{Parameters}} & \multicolumn{4}{|c|}{\textbf{Performance}} \\ \cline{5-8}  
& & & & \textbf{R1} & \textbf{R2} & \textbf{R3} & \textbf{R4} \\ 
\hline 
\textbf{AM} & Average Majority & \cite{Gao2007} & -- & 3.07 & 41.8 & 29.5 & 34.1 \\ 
\hline
\textbf{Ada} & Adaboost & \cite{Freund1997} & -- & 5.25 & 41.1 & 57.5 & \textbf{19.7} \\
\hline
\textbf{OnAda} & Fan's Online Adaboost & \cite{Fan1999} & Window size: $W=100$ & 2.25 & 41.9 & 39.3 & 19.8 \\
\hline
\textbf{Wang}  & Wang's Online Adaboost & \cite{Wang2003} & Window size: $W=100$ & 1.73 & 40.5 & 32.7 & 19.8 \\
\hline
\textbf{DDD} & Diversity for Dealing with Drifts & \cite{Minku2012} & Diversity parameters: $\lambda_l=1$, $\lambda_h=0.1$ & 0.72 & 39.7 & 24.6 & 20.0 \\ 
\hline
\textbf{WM} & Weighted Majority algorithm & \cite{Littlestone1994} & Multiplicative parameter: $\beta=0.5$ & 0.29 & 22.9 & 14.1 & 67.4 \\ 
\hline
\textbf{Blum} & Blum's variant of WM & \cite{Blum1997} & Multiplicative parameters: $\beta=0.5$, $\gamma=1.5$ & 1.64 & 37.3 & 22.6 & 68.1 \\ 
\hline
\textbf{TrackExp} & Herbster's variant of WM & \cite{Herbster1998} & Multiplicative and sharing parameters: $\beta=0.5$, $\alpha=0.25$ & 0.52 & 23.1 & 14.8 & 22.0 \\ 
\hline
\textbf{PWM} & \textbf{Perceptron Weighted Majority} & our work & -- & \textbf{0.19} & \textbf{14.3} & \textbf{4.1} & \textbf{31.5}\\ 
\hline
\end{tabular}
\end{center}
\label{tab:schemes}
\end{table*}

For each data set we consider a set of $8$ learners and we use logistic regression classifiers 
for the learners' local predictions.
Each local classifier is pre--trained using an individual training data set 
and kept fixed for the whole simulation (except for the OnAda, Wang, and DDD schemes, in which the base classifiers are retrained online).
The training and testing procedures are as follow. 
From the whole data set we select $8$ training data sets, each of them consisting of $Z$ sequential records.
$Z$ is equal to $5,000$ for the data sets \textbf{R1} and \textbf{R3}, and $2,000$ for \textbf{R2} and \textbf{R4}. 
Then we take other sequential records ($20,000$ for \textbf{R1} and \textbf{R3}, and $8,000$ for \textbf{R2} and \textbf{R4}) 
to generate a set in which the local classifiers are tested, and the results are used to train offline Adaboost.
Finally, we select other sequential records ($20,000$ for \textbf{R1} and \textbf{R3}, $21,000$ for \textbf{R2}, and $26,000$ for \textbf{R4}) 
to generate the testing set that is used to run the simulations and test all the considered schemes.

Table \ref{tab:schemes} reports the final mis--classification probability in percentages (i.e., multiplied by $100$) obtained for each data set for the considered schemes. 
For the first three data sets, which exhibit concepts drifts, 
the schemes that update their models after each instance (DDD, WM, Blum, TrackExp, and PWM) outperform the static schemes (AM and Ada) and the schemes that update their model after a chunk of instances enters the system (OnAda and Wang).
This result shows that the static schemes are not able to adapt to changes in concept, and the schemes that need to wait for a chunk of data adapt slowly because 1) they have to wait for the last instance of the chuck before updating the model, and 2) a chuck of data can contain instances belonging to different concepts, hence the model built on it can be inaccurate to predict the current concept. 

Importantly, in the first three data sets PWM outperforms all the other schemes, whereas the second best scheme is WM. 
The gain of PWM (in terms of reduction of the mis--classification probability) with respect to WM is about $34\%$ for \textbf{R1}, $38\%$ for \textbf{R2}, and $71\%$ for \textbf{R3}.
We remark that the main differences among our scheme and WM are 1) the weights update rule (additive vs. multiplicative), and 2) the weight $w_{i0}^{(n)}$ associated to the virtual learner that always sends the local prediction $1$.
To investigate the real reason of the gain of PWM we tested also a version of PWM in which $w_{i0}^{(n)}=0$, $\forall \, n$, obtaining the following percentage of mis--classifications in the first three data sets: $0.23$, $14.4$ and $4.1$.
Hence, the weight $w_{i0}^{(n)}$ can slightly help to increase the accuracy of the distributed system, but the main reason why PWM outperforms WM in these data sets is the update rule. 


Differently from the first three data sets, in \textbf{R4}, the data set that does not contain drifts, Ada, OnAda, and Wang outperform the other schemes. 
In fact, they exploit many stored labeled instances to build their models, and this results in more accurate models when the data are generated from a static distribution.

\subsection{Different Data Streams}
\label{sec:distr}

In this subsection we evaluate PWM using synthetic data sets in which different learners observe different data streams, and analyze the impact of delayed labels, missing labels, and asynchronous learners.
First, we shortly describe the data sets, then we discuss the results.

\subsubsection{Synthetic Data Sets}
\label{sec:sets}

We consider three synthetic data sets to carry on different experiments.
The first data set represents a separating hyperplane that rotates slowly, we use it to simulate gradual drifts \cite{Baena06, Minku2012, Zliobaite10}.  
Similar data sets are widely adopted in the stream mining literature dealing with concept drift \cite{Gao2007, Masud2009, Minku2012, Crammer10}. 
In the third data set, similarly to \cite{Willett00}, each learner observes a local event that is embedded in a zero--mean Gaussian noise.
Concept drifts occur because the accuracies of the observations evolve following Markov processes. 
The third data is a simple Gaussian distributed data set 
in which the concept is stable.
We use this data set because we can analytically compute the optimal mis--classification probability $P^O(\mathbf{D}_N)$ and investigate how strict the bound $\mathbf{B}(\mathbf{D}_N)$ is.

\textbf{S1: Rotating Hyperplane.} Each learner $i$ observes a $3$--dimensional instance $x_i^{(n)} = (x_{i,1}^{(n)}, x_{i,2}^{(n)}, x_{i,3}^{(n)})$ that is uniformly distributed in $[-1 \; 1]^3$, and is independent from $x_i^{(m)}$, $n \neq m$, and from $x_j^{(m)}$, $i \neq j$. 
The label is a deterministic function of the instances observed by the first $\underline{K} < K$ learners (the other $K - \underline{K}$ learners observe irrelevant instances). 
Specifically, $y^{(n)} = 1$ if $\sum_{i=1}^{\underline{K}} \sum_{\ell=1}^3 \theta_{i, \ell}^{(n)} x_{i,\ell}^{(n)} \geq 0$,  $y^{(n)} = 0$ otherwise. The parameters $\theta_{i, \ell}^{(n)}$ are unknown and time--varying. As in \cite{Crammer10}, each $\theta_{i, \ell}^{(1)}$ is independently generated according to a zero-mean unit-variance Gaussian distribution $\mathcal{N}(0, 1)$, and $\theta_{i, \ell}^{(n)} = \theta_{i, \ell}^{(n-1)} + \delta_{i, \ell}^{(n)}$ where $ \delta_{i, \ell}^{(n)} \sim \mathcal{N}(0, 0.1)$. 

\textbf{S2: Distributed Event Detection.} Each learner $i$ monitors the occurrence of a particular local event.
Let $e_i^{(n)} \triangleq 1$ if the local event monitored by learner $i$ occurs at time instant $n$, $e_i^{(n)} \triangleq -1$ otherwise.
$e_i^{(n)}$ is an i.i.d. process, the probability that $e_i^{(n)} = 1$ is $0.05$, $\forall \, i, n$.
The observation of learner $i$ 
is $\mathbf{x}_i^{(n)} = e_i^{(n)} + \beta_i^{(n)}$, where $\beta_i^{(n)}$ is an i.i.d. zero--mean Gaussian process. 
To simulate concept drifts, we assume that a source can be in two different states: good or bad.
In the good state $\beta_i^{(n)} \sim \mathcal{N}(0, 0.5)$, in the bad state $\beta_i^{(n)} \sim \mathcal{N}(0, 1)$.
The state of the source evolves as a Markov process with a probability $0.01$ to transit from one state to the other.

\textbf{S3: Gaussian Distribution.} The labels are generated according to a Bernoulli process with parameter $0.5$, 
and the instance $\mathbf{x}^{(n)} = ( x_1^{(n)}, \ldots, x_K^{(n)} )$ is generated according to a $K$--dimensional Gaussian distribution $\mathbf{x}^{(n)} \sim \mathcal{N}\left(y^{(n)} \cdot \mu, \Sigma \right)$, where $\Sigma$ is the identity matrix.
That is, if the label is $1$ ($-1$) each component $x_i^{(n)}$ is independently generated according to a Gaussian distribution with mean $\mu$ ($-\mu$) and unitary variance.
A generic learner $i$ observes only the component $x_i^{(n)}$ of the whole instance $\mathbf{x}^{(n)}$.

\subsubsection{Results}
\label{sec:res1}

\begin{figure}
     \centering
          \includegraphics[width=\figw]{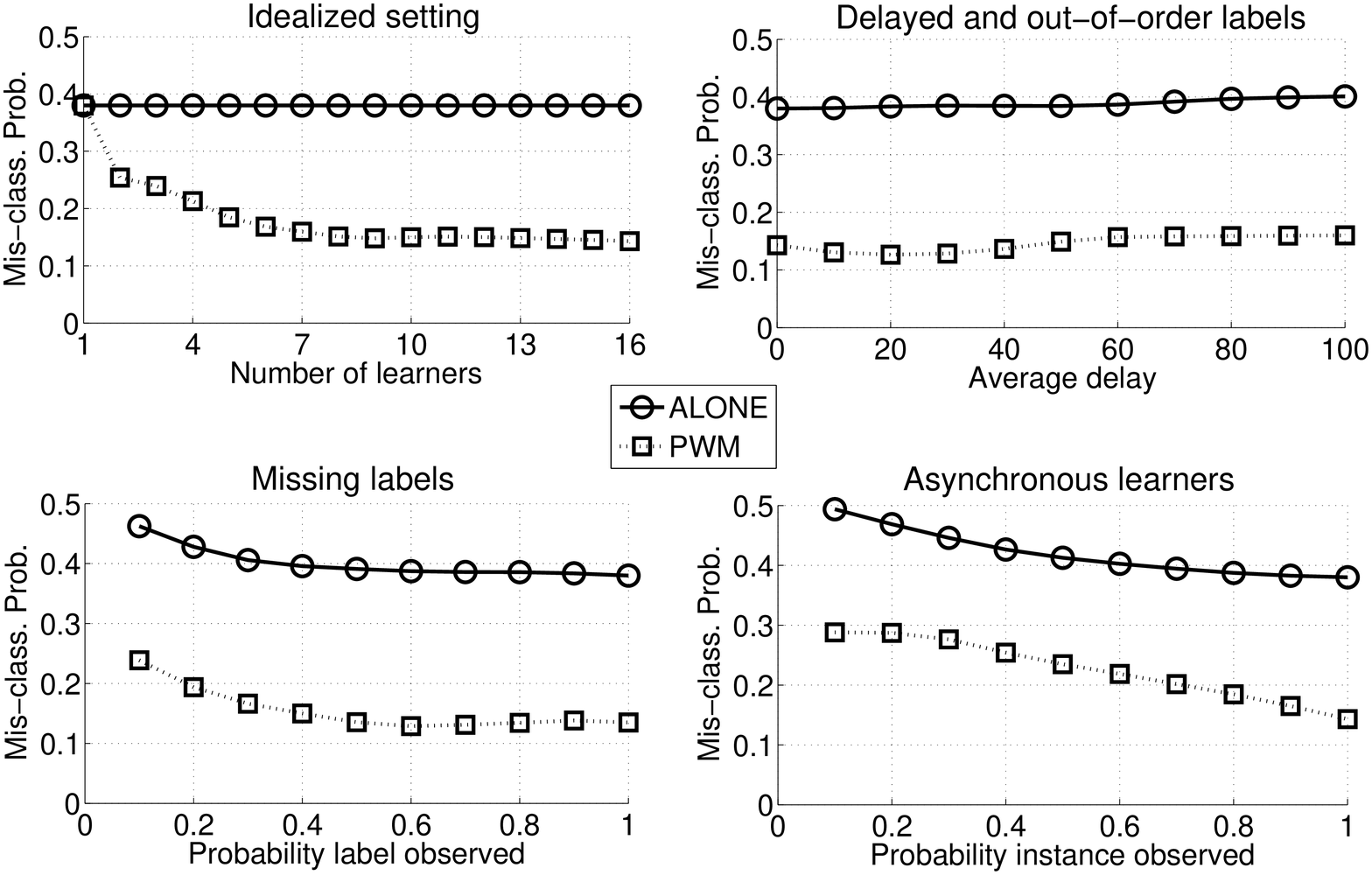}
\caption{Mis--classification probability of learner $1$ if it predicts alone and if it uses PWM, for the data set \textbf{S1}}
\label{fig:1}
\end{figure}

In the first set of experiments we adopt the synthetic data set \textbf{S1} to evaluate the mis--prediction probability of a generic learner, which we refer to as learner $1$, when it predicts by its own (ALONE), and when it adopts PWM.
We consider a set of $K=16$ learners, in which the last $8$ learners observe irrelevant instances.
For each simulation we generate a data set of $1,000$ instances.
We use non pre--trained online logistic regression classifiers 
for the learners' local predictions.
We run $1,000$ simulations and average the results.
The final results are reported in the four sub--figures of Fig. \ref{fig:1}, and are discussed in the following.

The top--left sub--figure shows how the mis--classification probability of learner $1$ varies, in the idealized setting (i.e., without the issues described in Section \ref{sec:dis}), with respect to the the number of learners that PWM aggregates. 
If there is only one learner, ALONE and PWM are equivalent, but the gap between the performance obtainable by ALONE and the performance achievable by PWM increases as the number of learners that PWM aggregates increases.
In particular, if the local predictions of all the learners are aggregated, the mis--classification probability of PWM is less than half the mis--classification probability of ALONE.
Notice that the performance of PWM remains constant from $8$ to $16$ learners, and this is a positive result because the last $8$ learners observe irrelevant instances. PWM automatically gives them a low weight such that their (noisy) local predictions do not influence the final prediction. 
In fact, the simulation for $K=16$ learners shows that the average absolute weight of the first $8$ learners is about twice the average absolute weight of the last $8$ learners.
In all the following experiments we consider $K=16$ learners.

Now we assume that learner $1$ observes the labels after some time instants, and each delay is uniformly distributed in $[0 \; D]$. 
The top--right sub--figure shows how the mis--classification probability varies with respect to the average delay $\frac{D}{2}$. 
We can see that the delay does not affect considerably the performance, in fact both mis--classification probabilities slightly increases if the delay increases and the gap between them remain constant. 
 
In the next experiment we analyze the impact of missing labels on the performance of learner $1$.
The bottom--left sub--figure shows how the mis--classification probability varies with respect to the probability that learner $1$ observes a label.
Even when the probability of observing a label is $0.1$, the mis--classification probability of PWM is about half the mis--classification probability of ALONE.
This gain is possible because learner $1$, adopting PWM, automatically exploits the fact that the other learners are learning. 

Similar considerations are valid when learner $1$ observes an instance with a certain probability (see the bottom--right sub--figure), which can be interpreted as the reciprocal of the arrival rate. 
The impact on the mis--classification probabilities of missing instances is stronger (i.e., the mis--classification probabilities are higher) than the impact of missing labels. 
In fact, when instances are not observed, not only learner $1$ does not update the weight vector, it also waits more time between two consecutive predictions, hence the concept between two consecutive predictions can change consistently. 
When the probability of observing an instance is $0.1$, the gain of PWM, with respect to ALONE, is about $40\%$.

\begin{figure}
     \centering
          \includegraphics[width=\figw]{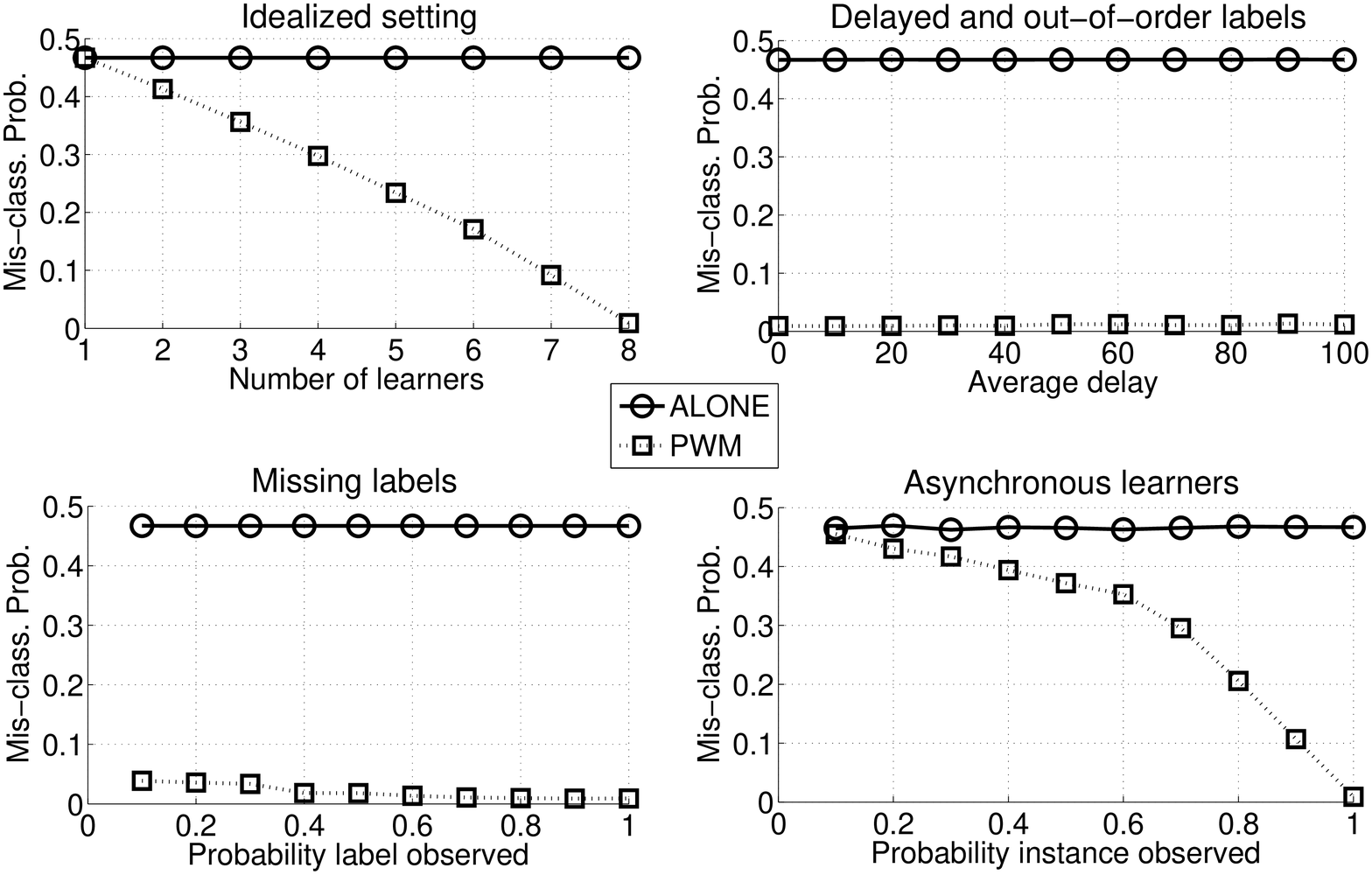}
\caption{Mis--classification probability of learner $1$ if it predicts alone and if it uses PWM, for the data set \textbf{S2}}
\label{fig:2}
\end{figure}

In the second set of experiments we use a similar set--up as in the first set of experiments, but we adopt the synthetic data set \textbf{S2}. 
We consider a set of $K=8$ learners and for each simulation we generate $10,000$ instances. 
Each learner uses a non pre--trained online logistic regression classifiers to learn the best threshold to adopt to classify the local event. 
We run $100$ simulations and average the results.
The final results are reported in the four sub--figures of Fig. \ref{fig:2}, and are briefly discussed in the following.
The top--left sub--figure shows that the mis--classification probability of PWM decreases linearly in the number of learners until the local prediction of all learners are aggregated, in this case the mis--classification probability of PWM is about $0.01$, whereas the mis--classification probability of ALONE is about $0.47$.
As in the first set of experiments, the delay does not affect the performance of the two schemes, and the performance of PWM is much better than the performance of ALONE even when the probability of observing the label is very low.
Differently from the first set of experiments, with the data set \textbf{S2} the performance of PWM is strongly affected by the synchronicity of the learners, and when the learners observe few instances the mis--classification of PWM becomes close to the mis--classification of ALONE.

In the last experiment we adopt the data set \textbf{S3} to investigate how strict the bound $\mathbf{B}(\mathbf{D}_N)$ is.
For each simulation we consider $K=8$ learners and generate a data set of $1,000$ instances.
We assume that the local prediction of learner $i$ is $-1$ if its observation $x_i^{(n)}$ is negative, $1$ otherwise. 
It is possible to show that, given the structure of the problem, this represents the most accurate policy for the local prediction, and the best possible aggregation rule is the average majority (AM). 
We run $10,000$ simulations and average the results.

\begin{figure}
     \centering
          \includegraphics[width=\figw]{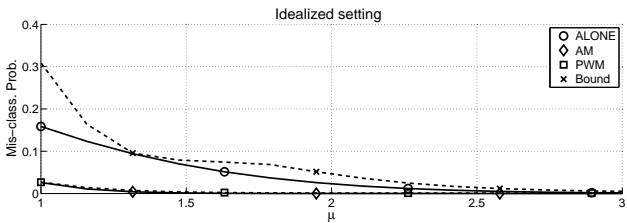}
\caption{The bound $\mathbf{B}(\mathbf{D}_N)$ and the mis--classification probability of learner $1$ if 1) it predicts by its own, 2) it uses AM, 3) it uses PWM, for the data set \textbf{S3}}
\label{fig:bound}
\end{figure}

Fig. \ref{fig:bound} shows the bound $\mathbf{B}(\mathbf{D}_N)$ and the mis--classification probability of learner $1$ if 1) it predicts by its own (ALONE), 2) it uses AM, and 3) it uses PWM, varying the parameter $\mu$.
If $\mu$ is low the instances corresponding to negative and positive labels are similar, hence it is more difficult to predict correctly the labels.
Fig. \ref{fig:bound} shows that, in this case, the mis--classification probability of PWM is much lower than the bound, and it is very close to the mis--classification probability of $AM$, that is the best aggregation rule in this scenario.
With the increase of $\mu$, the mis--classification probabilities of all the schemes decrease, and the bound become stricter to the real performance of PWM.

Notice that the curve representing the bound has a cusp at about $\mu = 1.75$. 
In fact, before this value $\mathbf{B}_1(\mathbf{D}_N)$ is stricter than $\mathbf{B}_2(\mathbf{D}_N)$, whereas for $\mu > 1.75$ $\mathbf{B}_2(\mathbf{D}_N)$ is lower than $\mathbf{B}_1(\mathbf{D}_N)$.
This agrees with Remark \ref{rem:8}: when $\mu$ is low the local classifiers are inaccurate (see ALONE), but their ensemble can be very accurate (see AM), and $\mathbf{B}_1(\mathbf{D}_N)$ is stricter than $\mathbf{B}_2(\mathbf{D}_N)$; whereas, when $\mu$ is high the local classifiers are very accurate and $\mathbf{B}_2(\mathbf{D}_N)$ becomes stricter than $\mathbf{B}_1(\mathbf{D}_N)$.

\section{Conclusion}
\label{sec:con}

We proposed a distributed online ensemble learning algorithm to classify data captured from distributed, heterogeneous, and dynamic data sources.
Our approach requires limited communication, computational, energy, and memory requirements.
We rigorously determined a bound for the worst--case mis--classification probability of our algorithm which depends on the mis--classification probabilities of the best static aggregation rule, and of the best local classifier. 
Importantly, this bound tends asymptotically to $0$ if the mis--classification probability of the best static aggregation rule tends to $0$.
We extended our algorithm and the corresponding bounds such that they can address challenges specific to the distributed implementation.
Simulation results show the efficacy of the proposed approach.
When applied to real data sets widely used by the literature dealing with dynamic data streams and concept drift, our scheme exhibits performance gains ranging from $34\%$ to $71\%$ with respect to state--of--the--art solutions.

\appendices 

\section{Proof of Lemma \ref{lem:1}} \label{app:1}
\begin{IEEEproof}
Since $P^{PWM}(\mathbf{D}_N) = P_i^{PWM}(\mathbf{D}_N)$, $\forall \, i$, we can derive the bound with respect to the mis--classification probability $P_i^{PWM}(\mathbf{D}_N)$ of a generic learner $i$.

The proof departs from \cite[Theorem 2]{Shalev2005}, which states that, for a general Perceptron algorithm (i.e., $s_i^{(n)}$ can belong to whatever subset of $\Re$),
if $\norm{\mathbf{s}^{(n)}} \leq R$, $\forall \, n$, then for every $\gamma > 0$ and vector $\mathbf{u} \in \Re^{K+1} $, $\norm{\mathbf{u}} = 1$, the number of prediction errors $N_e^{PWM}(\mathbf{D}_N) $ of the online Perceptron algorithm on the sequence $\mathbf{D}_N$ is bounded by
\be
N_e^{PWM}(\mathbf{D}_N)  \leq \left( \dfrac{R + \sqrt{\gamma D}}{\gamma} \right)^2
\label{eq:freund}
\ee
where $D = \sum_{n=1}^m d_n $, $d_n = \max \left( 0 , \gamma -  y^{n} ( \mathbf{u} \cdot \mathbf{s}^{(n)} )\right)$.
Starting from this bound, we exploit the structure of our problem (i.e., $s_i^{(n)} \in \{-1, 1 \}$) to derive the bound $\mathbf{B}_1(\mathbf{D}_N)$.

Since in our case $\norm{\mathbf{s}^{(n)}} = \sqrt{K+1}$, we can consider $R = \sqrt{K+1}$.
Notice that the last $K$ elements of $\mathbf{s}^{(n)}$, i.e., the local predictions,
represent a particular vertex of an hypercube in $\Re^{K}$, and the optimal a posteriori weight vector $\mathbf{w}^O$ represents an hyperplane in $\Re^{K}$ which separates the $2^K$ vertexes of the hypercube in two subsets $\mathcal{V}_{-1}$ and $\mathcal{V}_1$, representing the vertexes resulting in a negative and positive prediction respectively.
Now we consider two scenarios: (1) either $\mathcal{V}_{-1}$ or $\mathcal{V}_1$ are empty; (2) both $\mathcal{V}_{-1}$ and $\mathcal{V}_1$ are not empty.

We consider the first scenario.
In this situation the optimal policy $\mathbf{w}^O$ predicts always $-1$ or $1$, independently of the local predictions (this case is not very interesting in practice, but we analyze it for completeness).
The geometric interpretation is that the separating hyperplane does not intersect the hypercube.
Let $\gamma$ be the distance between the separating hyperplane and the closest vertex of the hypercube, and $\mathbf{u} = \frac{\mathbf{w}^O}{\norm{\mathbf{w}^O}}$.
If $\mathbf{w}^O$ predicts correctly the $n$-th instance, then $y^{n} ( \mathbf{u} \cdot \mathbf{s}^{(n)} ) \geq \gamma$, hence $d_n = 0$.
If $\mathbf{w}^O$ makes a mistakes in the $n$-th instance, then $y^{n} ( \mathbf{u} \cdot \mathbf{s}^{(n)} ) \leq -\gamma$ and $y^{n} ( \mathbf{u} \cdot \mathbf{s}^{(n)} ) \geq -\gamma - 2 \sqrt{K}$ (because the closest vertex is $2 \sqrt{K}$ distant from the farthest one), therefore $d_n \leq 2 \gamma + 2 \sqrt{K}$.
Hence, we obtain $D \leq 2 N^O(\mathbf{D}_N)  \left( \gamma + \sqrt{K}  \right) $, where $N_e^O(\mathbf{D}_N) $ is the number of mistakes made adopting $\mathbf{w}^O$, and
\be
\left( \dfrac{R + \sqrt{\gamma D}}{\gamma} \right)^2 \leq \left( \dfrac{ \sqrt{K + 1} + \sqrt{ 2 N_e^O(\mathbf{D}_N)  \gamma \left( \gamma + \sqrt{K} \right)}} { \gamma }  \right)^2 \nonumber
\ee
The right side of the above inequality is decreasing in $\gamma$.
Since we can consider other optimal a posteriori weight vectors $\mathbf{w}^O$ and since there is no constraint on how far the separating hyperplane could be with respect to the hypercube, taking the limit for $\gamma \to + \infty$ and dividing everything by $N$ we finally obtain
\be
P_{i}^{PWM}(\mathbf{D}_N)  \leq \dfrac{1}{N} \lim_{\gamma \to + \infty} \left( \dfrac{R + \sqrt{\gamma D}}{\gamma} \right)^2 \leq 2 P^O(\mathbf{D}_N)   \;\; , \nonumber
\ee
which is compatible with $P^{PWM}(\mathbf{D}_N) \leq \mathbf{B}_1(\mathbf{D}_N)$.

Now we consider the second scenario.
In this case the separating hyperplane intersects the hypercube.
Among all the optimal a posteriori weight vectors $\mathbf{w}^O$, we want to consider the one which separates the vertexes in $\mathcal{V}_{-1}$ from the vertexes in $\mathcal{V}_1$ with the largest margin possible in order to find the strictest bound defined by (\ref{eq:freund}).
However, since the bound must be valid for every linearly separable sets of vertexes $\mathcal{V}_{-1}$ and $\mathcal{V}_1$, we have to consider the worst case possible (in term of separating margin) with respect to the sets $\mathcal{V}_{-1}$ and $\mathcal{V}_1$.
It is easy to see that the worst case corresponds to the situation in which one vertex must be separated by all the vertexes it is connected with through an edge, and the best separating hyperplane must be equidistant from all these vertexes.
Since the distances between the vertexes and the separating hyperplane are invariant with respect to translation and rotation of both the hypercube and the hyperplane, we consider the hypercube with vertexes $\mathbf{v} = (v_1, \ldots, v_{K})$, $v_i \in {0, 2}$, and we want to find the hyperplane defined by the parameters $(a_0, \ldots, a_{K})$ which separates with the largest margin the vertex $\mathbf{v}^{(0)} = (0, \ldots, 0)$ from the vertexes $\mathbf{v}^{(i)} = (0, \ldots, 0, 2, 0, \ldots , 0)$ having $2$ in position $i$, $i = 1, \ldots, K$.
Imposing that the signed distance between $\mathbf{v}^{(0)} $ and the separating hyperplane is the opposite of the signed distance between $\mathbf{v}^{(i)} $ and the separating hyperplane, we obtain
\ba
\dfrac{a_0 + v_1^{(0)} a_1 + \ldots + v_{K}^{(0)} a_K}{\sqrt{\sum_{i+1}^K a_i^2 } }  &= - \dfrac{a_0 + v_1^{(i)} a_1 + \ldots + v_K^{(i)} a_K}{\sqrt{\sum_{i+1}^K a_i^2 } }  \nonumber\\
\rightarrow a_0 &= - a_i \nonumber
\end{align}
Repeating the same procedure for every vertex $\mathbf{v}^{(i)}$, $i = 1, \ldots, K$, we obtain that the best separating hyperplane must satisfy $a_0 = - a_i$, $\forall \, i$, hence the distance between it and $\mathbf{v}^{(0)} $  is $\frac{\abs{a_0}}{\sqrt{K a_0^2}} = \frac{1}{\sqrt{K}}$.
This means that we are always able to find an optimal a posterior weight vector $\mathbf{w}^O$ which separates the local prediction vectors which a margin of at least $\gamma = \frac{1}{\sqrt{K}}$.
Notice also that the maximum distance between the hyperplane defined by $\mathbf{w}^O$ and a vertex in the hypercube is $2\sqrt{K}-\frac{1}{\sqrt{K}}$.
Define $\gamma =  \frac{1}{\sqrt{K}}$ and $\mathbf{u} = \frac{\mathbf{w}^O}{\norm{\mathbf{w}^O}}$.
If $\mathbf{w}^O$ predicts correctly the $n$-th instance, then $y^{n} ( \mathbf{u} \cdot \mathbf{s}^{(n)} ) \geq \gamma$, hence $d_n = 0$.
If $\mathbf{w}^O$ makes a mistakes in the $n$-th instance, then $y^{n} ( \mathbf{u} \cdot \mathbf{s}^{(n)} ) \leq -\gamma$ and $y^{n} ( \mathbf{u} \cdot \mathbf{s}^{(n)} ) \geq -2\sqrt{K}+\frac{1}{\sqrt{K}}$, therefore $d_n \leq 2 \sqrt{K} $.
Finally, we obtain $D \leq 2 \sqrt{K} N_e^O(\mathbf{D}_N)  $ and
\be
P_{i}^{PWM}(\mathbf{D}_N)  \leq \dfrac{1}{N} \left( \dfrac{R + \sqrt{\gamma D}}{\gamma} \right)^2 \leq 2 K  P^O(\mathbf{D}_N)  + \dfrac{K (K + 1)}{N} \nonumber
\ee

\end{IEEEproof}

\section{Proof of Lemma \ref{lem:2}} \label{app:2}

\begin{IEEEproof}
Since $P^{PWM}(\mathbf{D}_N) = P_i^{PWM}(\mathbf{D}_N)$, $\forall \, i$, we can derive the bound with respect to the mis--classification probability $P_i^{PWM}(\mathbf{D}_N)$ of a generic learner $i$.

PWM updates its weight vector only on those instances in which it makes a mistake.
We denote with the superscript $n$ the parameters of the system during the $n$-th mistake.
We have
\ba
\norm{\mathbf{w}_i^{n+1}}^2 &=  \norm{   \mathbf{w}_i^{n} + y^{n} \mathbf{s}^{n}  }^2 =
\norm{   \mathbf{w}_i^{n}  }^2 + \norm{  \mathbf{s}_i^{n}  }^2 + 2 y^{n} \mathbf{w}_i^{n}  \cdot \mathbf{s}^{n} \nonumber\\
&\leq \norm{   \mathbf{w}_i^{n}  }^2 + \norm{  \mathbf{s}^{n}  }^2 = \norm{   \mathbf{w}_i^{n}  }^2 + K + 1
\nonumber
\end{align}
where the first inequality is valid because the system makes an error, hence $y^{n} \mathbf{w}_i^{n}  \cdot \mathbf{s}^{n} \leq 0$.
By applying a straightforward inductive argument we obtain
\be
\norm{\mathbf{w}_i^{n+1}}^2 \leq n (K + 1) \nonumber
\ee

To simplify the notations, we denote by $z$ the number of errors made by the most accurate classifier and by $v^*$ the number of most accurate classifiers, i.e., $z = N P^*(\mathbf{D}_N) $ and $v^* = v^*(\mathbf{D}_N) $
After the system makes $n$ errors, the weight associated to the most accurate classifiers is at least $n - 2z$ (it increases by one unit at least $n-z$ times, and decreases by one unit at most $z$ times).
Hence,
\be
\norm{\mathbf{w}_i^{n+1}}^2 \geq v^* \, (n - 2z)^2 \nonumber
\ee

Combining the two above inequalities we obtain
\be
v^* n^2 - ( 4v^* z + K + 1) n + 4 v^* z^2 \leq 0 \nonumber
\ee
which implies $n \leq 2 z + \frac{K + 1}{2 v^*} + \sqrt{\left(\frac{K + 1}{2 v^*}\right)^2 + \frac{2 (K + 1) x}{v^*}}$.
Dividing by $N$ we obtain $P^{PWM}(\mathbf{D}_N) \leq \mathbf{B}_2(\mathbf{D}_N)$.
\end{IEEEproof}

\section{Proof of Theorem \ref{teo:3}} \label{app:3}

\begin{IEEEproof}
We denote by $\mathbf{w}_{i,c}^{0}$ the weight vector of learner $i$ at the beginning of a generic concept $\mathbf{S}_c^{(n)} $, and use the superscript $n$ to denote the parameters of the system during the $n$-th mistake inside the concept $\mathbf{S}_c^{(n)} $.
We have
\ba
\norm{\mathbf{w}&_{i,c}^{n+1}}^2 =  \norm{   \mathbf{w}_{i,c}^{n} + y^{n} \mathbf{m}^{n}  }^2 =
\norm{   \mathbf{w}_{i,c}^{n}  }^2 + \norm{  \mathbf{m}^{n}  }^2 + \nonumber\\
&2 y^{n} \mathbf{w}_{i,c}^{n}  \cdot \mathbf{m}^{n} \leq \norm{ \mathbf{w}_{i,c}^{n}  }^2 + \norm{  \mathbf{m}^{n}  }^2 = \norm{   \mathbf{w}_{i,c}^{n}  }^2 + K + 1 \nonumber
\end{align}
where the first inequality is valid because the system makes an error, hence $y^{n} \mathbf{w}_{i,c}^{n}  \cdot \mathbf{m}^{n} \leq 0$.
By applying a straightforward inductive argument we obtain
\be
\norm{\mathbf{w}_{i,c}^{n+1}}^2 \leq \norm{\mathbf{w}_{i,c}^0}^2 + n (K + 1)
\label{eq:proof1}
\ee

Since the concept $\mathbf{S}_c^{(n)}$ is learnable, there exists a unit vector $\mathbf{u} \in \Re^{K+1} $, $\norm{\mathbf{u}} = 1$, and $\gamma > 0$ such that $y^{n} \mathbf{u} \cdot \mathbf{m}^{n} \geq \gamma$, for every labeled instances of the current concept.
We have
\ba
\mathbf{w}_{i,c}^{n+1} \cdot \mathbf{u} &= \mathbf{w}_{i,c}^{n} \cdot \mathbf{u} + y^{n} \mathbf{m}^{n} \cdot \mathbf{u} \geq \mathbf{w}_{i,c}^{n} \cdot \mathbf{u} + \gamma \nonumber
\end{align}

Hence
\ba
\mathbf{w}_{i,c}^{n+1} \cdot \mathbf{u} \geq \mathbf{w}_{i,c}^{0} \cdot \mathbf{u} + n \gamma
\label{eq:proof2}
\end{align}

Combining (\ref{eq:proof1}) with (\ref{eq:proof2}) we obtain
\ba
\sqrt{ \norm{\mathbf{w}_{i,c}^0}^2 + n (K + 1) } &\geq \norm{\mathbf{w}_{i,c}^{n+1}} \geq \mathbf{w}_{i,c}^{n+1} \cdot \mathbf{u} \geq \nonumber\\
&\geq \mathbf{w}_{i,c}^{0} \cdot \mathbf{u} + n \gamma \geq - \norm{\mathbf{w}_{i,c}^0} + n \gamma \nonumber
\end{align}

For $n \geq \frac{\norm{\mathbf{w}_{i,c}^0}}{\gamma}$ (if this is not valid then $n$ is bounded by $\frac{\norm{\mathbf{w}_{i,c}^0}}{\gamma}$ which stricter than the following bound) we obtain
\ba
& \norm{\mathbf{w}_{i,c}^0}^2 + n (K + 1) \geq \norm{\mathbf{w}_{i,c}^0}^2 - 2 \norm{\mathbf{w}_{i,c}^0} n \gamma + \gamma^2 n^2 \nonumber\\
&\rightarrow n \leq \dfrac{2 \norm{\mathbf{w}_{i,c}^0} \gamma + K + 1}{\gamma^2}
\nonumber
\end{align}

As shown in Appendix \ref{app:1}, we can take $\gamma = \frac{1}{\sqrt{K}}$.
Hence, we obtain
\ba
n \leq 2 \sqrt{K} \norm{\mathbf{w}_{i,c}^0} + K (K + 1)
\label{eq:fin}
\end{align}

In the first concept $\mathbf{w}_{i,1}^{0} $ is initialized to $0$, thus the number of errors at the end of the first concept is upper--bounded by a  bounded function of $K$, and in turns also the norm of the weight vector $\mathbf{w}_{i,2}$ at the beginning of the second concept is bounded by a function of $K$.
Exploiting (\ref{eq:fin}) and using an inductive argument we can conclude that the number of errors at the end of each concept is upper bounded by a bounded function $K$. 
Since there are a finite number of concepts, the total number of errors is upper-bounded by a bounded function of $K$.
Finally, dividing it by $N$, we obtain $P_{i}^{PWM}(\mathbf{D}_N) \to 0$, that implies $P^{PWM}(\mathbf{D}_N) \to 0$.
\end{IEEEproof}

\section{Proof of Theorem \ref{teo:4}} \label{app:4}

\begin{IEEEproof}
Denote by $\overline{y}^{(n)}$ the $n$-th label observed by learner $i$, and by $\overline{\textbf{x}}^{(n)}$ the corresponding instance. 
Let $\overline{\mathbf{D}}_M = \left( ( \overline{\textbf{x}}^{(1)}, \overline{y}^{(1)}), \ldots , ( \overline{\textbf{x}}_{\ell}^{(1)} , \overline{y}^{(1)}),  \ldots, ( \overline{\textbf{x}}_{\ell}^{(M)}, \overline{y}^{(M)}) \right)$ the sequence of the $M$ labeled instances observed by learner $i$ until time instant $N$.
Notice that $N-\overline{d}_i \leq M \leq N$.
We can applied Theorem \ref{teo:1} to $\overline{\mathbf{D}}_M$ (the bound of Theorem \ref{teo:1} is valid also for the mis--classification probability of a generic learner $i$), obtaining
\be
P_i^{PWM}(\overline{\mathbf{D}}_M) \leq \mathbf{B}(\overline{\mathbf{D}}_M)  \nonumber
\ee
$\overline{\mathbf{D}}_M$ is a permutation of a subset of $\mathbf{D}_N$, hence the number of errors made by the optimal aggregation rule and by the best classifier in $\overline{\mathbf{D}}_M$ cannot by higher than those made in $\mathbf{D}_N$, i.e., $P^O(\overline{\mathbf{D}}_M) M \leq P^O(\mathbf{D}_N) N$ and $P^*(\overline{\mathbf{D}}_M) M \leq P^*(\mathbf{D}_N) N$. 
The number of errors learner $i$ makes over $\mathbf{D}_N$ adopting the PWM scheme are equal to the number of errors learner $i$ makes over $\overline{\mathbf{D}}_M$ plus the number of errors it makes over the label instances whose labels are not observed. 
Since the last term is bounded by $\overline{d}_i$, we obtain 
\ba
P_i^{PWM}(\mathbf{D}_N) &\leq \dfrac{1}{N} P_i^{PWM}(\overline{\mathbf{D}}_M) M + \dfrac{\overline{d}_i}{N} \leq \mathbf{B}(\mathbf{D}_N) + \dfrac{\overline{d}_i}{K} \nonumber 
\end{align}
By applying (\ref{eq:P_sys}) we we conclude the proof.
\end{IEEEproof}

\section{Proof of Theorem \ref{teo:5}} \label{app:4bis}

\begin{IEEEproof}
The proof applies the same is methodology as the proof of Theorem \ref{teo:4}. 
The number of errors a generic learner $i$ makes in each concepts can be divided into two contributions.
The first contribution represents the number of errors $i$ makes over the sequence of labeled instances it observes. 
Theorem \ref{teo:3} proves that such a term tends to $0$.
The second contribution represents the number of errors learner $i$ makes over the label instances whose labels are not observed. 
Such a term is bounded by $\frac{\overline{d}_i}{N}$, that tends to $0$.
\end{IEEEproof}

\section{Proof of Theorem \ref{teo:6}} \label{app:5}

\begin{IEEEproof}
Inside this proof, to simplify the notations, we denote by $n = P^{PWM}(\mathbf{D}_N) N$ the number of errors made by PWM and by $t = N_e^{PWM}$ the number of observed errors.
The number of observed errors $t$ is a binomial with parameters $n$ and $\mu$.
Exploiting a Chernoff--Hoeffding inequality \cite{Boucheron04} we can write
\be
P \left[ t \leq n (\mu - \gamma ) \right] \leq e^{-2 \gamma^2 n } \leq e^{-2 \gamma^2 t} \nonumber
\ee
which implies that
\be
P \left[ n \geq \dfrac{t}{\mu - \gamma} \right] \leq e^{-2 \gamma^2 t} \nonumber
\ee

Since PWM updates its weight vector only on those instances on which an error is observed, the bound $\mathbf{B}(\mathbf{D}_N)$ for non perfectly observable labels must be interpreted as bounds for the number of observed errors, i.e.,
\be
\dfrac{t}{N} \leq \mathbf{B}(\mathbf{D}_N) \nonumber
\ee
The two inequalities above, with the change of variable $e^{-2 \gamma^2 t} = \epsilon$,
imply (\ref{eq:boundR5})
with probability at least $1 - \epsilon$.
\end{IEEEproof}

\section{Proof of Theorem \ref{teo:7}} \label{app:5bis}

\begin{IEEEproof}
If the number of errors $\overline{N}_{e,c}^{PWM}$ PWM makes in concept $c$ is finite, then $\frac{\overline{N}_{e,c}^{PWM}}{N} \to 0$. 
If $\overline{N}_{e,c}^{PWM}$ is unbounded, then by  a Chernoff--Hoeffding bound \cite{Boucheron04} we have $N_{e,c}^{PWM} = \mu \cdot \overline{N}_{e,c}^{PWM}$ with probability $1$.
Using the same arguments as in the proof of Theorem \ref{teo:3}, we can say that $\frac{N_{e,c}^{PWM}}{N} \to 0$, hence $\frac{\overline{N}_{e,c}^{PWM}}{N} \to 0$.
Therefore, $P^{PWM}(\mathbf{D}_N) = \frac{\sum_c \overline{N}_{e,c}^{PWM}}{N} \to 0$.
\end{IEEEproof}

\section{Proof of Theorem \ref{teo:10}} \label{app:7}

\begin{IEEEproof}
Denote by $\overline{\mathbf{D}}_M$ the subset of labeled instances of $\mathbf{D}_N$ in which all the learners are synchronized.
The number of errors learner $i$ makes over $\mathbf{D}_{N}$ adopting the PWM scheme is equal to the number of errors learner $i$ makes over $\overline{\mathbf{D}}_M$, plus the number of errors it makes over the $N - M$ label instances in which some learners do not observe the instances.
The weight vector $\mathbf{w}_{i,s}^{(n)}$ is used only to predict the instances in $\overline{\mathbf{D}}_M$, and is updated only in these instances.
Therefore, we can applied Theorem \ref{teo:1} to $\overline{\mathbf{D}}_M$ (the bound of Theorem \ref{teo:1} is valid also for the mis--classification probability of a generic learner $i$), obtaining
\be
P_i^{PWM}(\overline{\mathbf{D}}_M) \leq \mathbf{B}(\overline{\mathbf{D}}_M) \nonumber
\ee
$\overline{\mathbf{D}}_M$ is a subset of $\mathbf{D}_N$, hence the number of errors made by the optimal aggregation rule and by the best classifier in $\overline{\mathbf{D}}_M$ cannot by higher than those made in $\mathbf{D}_N$, i.e., $P^O(\overline{\mathbf{D}}_M) M \leq P^O(\mathbf{D}_N) N$ and $P^*(\overline{\mathbf{D}}_M) M \leq P^*(\mathbf{D}_N) N$. 
Therefore the number of errors made by $i$ in $\overline{\mathbf{D}}_M$ is $P_i^{PWM}(\overline{\mathbf{D}}_M)  M \leq \mathbf{B}(\overline{\mathbf{D}}_M) M \leq \mathbf{B}(\mathbf{D}_N) N$, which implies that the contribution of $i$ to $P^{PWM}(\mathbf{D}_N)$ is at most
\be
\mathbf{B}(\mathbf{D}_N) + \alpha  \nonumber
\ee
The proof is concluded by summing the contributions of all learners and dividing the result by $K$.
\end{IEEEproof}

\section{Proof of Theorem \ref{teo:11}} \label{app:7bis}

\begin{IEEEproof}
Using the notations and considerations of the proof of Theorem \ref{teo:10} we can state that, for a generic learner, the mis--classification probability in the sequence $\overline{\mathbf{D}}_M$ tends to $0$ (because we can apply Theorem \ref{teo:3}), whereas the mis--classification probability over the instances in which some learners do not observe the instances is bounded by $\alpha$.
\end{IEEEproof}

\bibliographystyle{IEEEtran}
\bibliography{IEEEabrv,biblio}

\end{document}